\documentclass[letterpaper,twocolumn,10pt]{article}
\usepackage{usenix2019_v3}

\usepackage{booktabs} 
\usepackage[english]{babel}
\usepackage{xcolor}
\usepackage{xspace}
\usepackage{colortbl}
\usepackage{pifont}
\usepackage{subfigure}
\usepackage{graphicx}
\usepackage[inline]{enumitem}
\usepackage{booktabs}
\usepackage{threeparttable}
\usepackage{times}
\usepackage[small, compact]{titlesec}
\usepackage[textfont={it}, font={small,bf}, skip=1pt]{caption}

\usepackage{multirow}
\usepackage{fancybox} 
\usepackage{amsmath}
\usepackage{hhline}
\usepackage{amsthm}
\usepackage{comment}
\usepackage{tabularx}
\usepackage{bbm}
\usepackage{amsfonts}
\usepackage{pythonhighlight}

\usepackage{hyperref}
\hypersetup{
  colorlinks=true,      
  linkcolor=blue,       
  citecolor=magenta,    
  filecolor=cyan,       
  urlcolor=red          
}

\newtheorem{lemma}{Lemma}
\usepackage{titling}
\date{}

\definecolor{boxcolor}{gray}{0.9}

\newtheorem{corollary}{Corollary}[lemma]

\usepackage[linesnumbered, noend, noline]{algorithm2e}
\usepackage{float}
\usepackage{algorithmicx}

\SetCommentSty{mycommfont}
\newcommand{\algcomment}[1]{\footnotesize{\Comment{#1}}}
\newcommand{\algrule}[1][.7pt]{\par\vskip.5\baselineskip\hrule height #1\par\vskip.5\baselineskip}

\newenvironment{denseitemize}{
\begin{itemize}[topsep=2pt, partopsep=0pt, leftmargin=1.5em]
  \setlength{\itemsep}{2pt}
  \setlength{\parskip}{0pt}
  \setlength{\parsep}{0pt}
}{\end{itemize}}

\newenvironment{denseenum}{
\begin{enumerate}[topsep=2pt, partopsep=0pt, leftmargin=1.5em]
  \setlength{\itemsep}{2pt}
  \setlength{\parskip}{0pt}
  \setlength{\parsep}{0pt}
}{\end{enumerate}}

\definecolor{codegreen}{rgb}{0,0.6,0}
\definecolor{codegray}{rgb}{0.5,0.5,0.5}
\definecolor{codepurple}{rgb}{0.58,0,0.82}
\definecolor{backcolour}{rgb}{0.95,0.95,0.92}

\lstdefinestyle{codestyle}{
    commentstyle=\color{codegreen},
    keywordstyle=\color{black}\bfseries,
    numberstyle=\color{codegray},
    stringstyle=\color{codepurple},
    basicstyle=\ttfamily\mdseries\scriptsize,
    emph={selector,Oort},
    emphstyle={\color{magenta}\bfseries},
    breakatwhitespace=false, 
    frame=bottomline,        
    breaklines=true,                 
    captionpos=b,                    
    keepspaces=true,                 
    numbers=left,                    
    numbersep=5pt,                  
    showspaces=false,                
    showstringspaces=false,
    showtabs=false,                  
    tabsize=2
}
\def\ie{{i.e.}}
\def\eg{{e.g.}}

\def\name{Oort\xspace}
\usepackage{tikz}
\newcommand*\circled[1]{\tikz[baseline=(char.base)]{
            \node[shape=circle,draw,inner sep=0pt] (char) {#1};}}
\algnewcommand{\LeftComment}[1]{\Statex \(\triangleright\) #1}

\newcommand{\update}[1]{\textcolor{black}{#1}}

\begin{document}
\title{\fontsize{16}{16}{\textbf{\name: Efficient Federated Learning via Guided Participant Selection}}}

\author{
\rm Fan Lai, Xiangfeng Zhu, Harsha V. Madhyastha, Mosharaf Chowdhury
\\
\itshape{University of Michigan}
}
  
\maketitle

\begin{abstract}
Federated Learning (FL) is an emerging direction in distributed machine learning (ML) that enables in-situ model training and testing on edge data.
Despite having the same end goals as traditional ML, FL executions differ significantly in scale, spanning thousands to millions of participating devices. 
As a result, data characteristics and device capabilities vary widely across clients. 
Yet, existing efforts randomly select FL participants, which leads to poor model and system efficiency. 

In this paper, we propose \name to improve the performance of federated training and testing with guided participant selection. 
With an aim to improve time-to-accuracy performance in model training, \name
prioritizes the use of those clients who have both data that offers the
greatest utility in improving model accuracy and the capability to run
training quickly.
To enable FL developers to interpret their results in model testing, 
\name enforces their requirements on the distribution of participant data while improving the duration of federated testing by cherry-picking clients. 
Our evaluation shows that, compared to existing participant selection mechanisms, 
\name improves time-to-accuracy performance by 1.2$\times$-14.1$\times$ and final model accuracy by 1.3\%-9.8\%, while efficiently enforcing developer-specified model testing criteria at the scale of millions of clients.
\end{abstract}

\section{Introduction} 

Machine learning (ML) today is experiencing a paradigm shift from cloud datacenters toward the edge \cite{federated-learning, fl-survey}. 
Edge devices, ranging from smartphones and laptops to enterprise surveillance cameras and edge clusters, routinely store application data and provide the foundation for machine learning beyond datacenters. 
With the goal of not exposing raw data, large companies such as Google and Apple deploy \emph{federated learning (FL)} for computer vision (CV) and natural language processing (NLP) tasks across user devices\cite{apple-dp, ggkeyboard, fate, firefox}; NVIDIA applies FL to create medical imaging AI \cite{nvidia-fl}; smart cities perform in-situ image training and testing on AI cameras to avoid expensive data migration \cite{focus, camera-cluster, fed-oj}; and video streaming and networking communities use FL to interpret and react to network conditions \cite{puffer-web, puffer}.

Although the life cycle of an FL model is similar to that in traditional ML, the underlying execution in FL is spread across thousands to millions of devices in the wild. 
Similar to traditional ML, the FL developer often first prototypes model architectures and hyperparameters 
with a proxy dataset.
After selecting a suitable configuration, she can use federated training to improve model performance by training across a crowd of participants \cite{federated-learning, fl-survey}. 
The wall clock time for training a model to reach an accuracy target (\ie, time-to-accuracy) is still a key performance objective even though it may take significantly longer than centralized training \cite{fl-survey}. 
To circumvent biased or stale proxy data in hyperparameter tuning \cite{fl-distri}, to inspect these models being trained, or to validate deployed models after training \cite{puffer, mobile-mlapps}, developers may want to perform federated testing on the real-life client data, wherein enforcing their requirements on the testing set (\eg, $N$ samples for each category or following the representative categorical distribution\footnote{A categorical distribution is a discrete probability distribution showing how a random variable can take the  result from one of $K$ possible categories.}) 
is crucial for them to reason about model performance under different data characteristics \cite{data-val, fl-distri}.


Unfortunately, clients may not all be simultaneously available for FL training or testing \cite{fl-survey}; they may have heterogeneous data distributions and system capabilities \cite{noniid,federated-learning}; and including too many may lead to wasted work and suboptimal performance \cite{federated-learning} (\S\ref{sec:background}). 
Consequently, a fundamental problem in practical FL is the \emph{selection of a ``good'' subset of clients as participants}, where each participant locally processes its own data, and only their results are collected and aggregated at a (logically) centralized coordinator.

Existing works optimize for \emph{statistical model efficiency} (\ie, better training accuracy with fewer training rounds) \cite{adapt-sgd, fl-heternet, fl-adver, keyboard-acl} or \emph{system efficiency} (\ie, shorter rounds) \cite{fedavg, fl-kmeans}, while randomly selecting participants. 
Although random participant selection is easy to deploy, unfortunately, it results in poor performance of federated training because of large heterogeneity in device speed and/or data characteristics.
Worse, random participant selection can lead to biased testing sets and loss of confidence in results. 
As a result, developers often resort to more participants than perhaps needed \cite{gg-keyeval, fl-distri}.

We present \name for FL developers to enable guided participant selection throughout the life cycle of an FL model (\S\ref{sec:overview}). 
Specifically, \name cherry-picks participants to improve time-to-accuracy performance for federated training, 
and it enables developers to specify testing criteria for federated model testing.  
It makes informed participant selection by relying on the information already available in existing FL solutions \cite{fl-survey} with little modification.

Selecting participants for federated training is challenging because of the trade-off between heterogeneous system and statistical model utilities both across clients and of any specific client over time (as the trained model changes). 
First, simply picking clients with high statistical utility can lead to longer training rounds due to the coupled nature of client data and system performance. 
The challenge is further exacerbated by the large population, as capturing the latest utility of all clients is impractical. 
As such, we identify clients with high statistical utility, which is measured in terms of their most recent aggregate training loss, adjusted for spatiotemporal variations, and penalize the utility of a client if her system speed is likely to elongate the duration necessary to complete global aggregation. 
To navigate the sweet point of jointly maximizing statistical and system efficiency, we adaptively allow for longer training rounds to admit clients with higher statistical utility. 
We then employ an online exploration-exploitation strategy to probabilistically select participants among high-utility clients for robustness to outliers. 
Our design can accommodate diverse selection criteria (\eg, fairness), and deliver improvements while respecting privacy (\S\ref{sec:model-training}).

Although FL developers often have well-defined requirements on their testing data, 
satisfying these requirements is not straightforward. 
Similar to traditional ML, developers may request a testing dataset that follows the 
global distribution to avoid testing on all clients \cite{fedvc, fl-distri}.
However, clients' data characteristics in some private FL scenarios may not be available \cite{diff-fl, ggkeyboard}. 
To preserve the deviation target of participant data from the global, 
\name performs participant selection by bounding the number of participants needed. 
Second, for cases where clients' data characteristics are provided \cite{fed-oj}, 
developers can specify specific distribution of the testing set to debug model efficiency 
 (\eg, using balanced distribution) \cite{inspect-fl, positive-fl}. 
At scale, satisfying this requirement in FL suffers large overhead. 
Therefore, we propose a scalable heuristic to efficiently enforce developer requirements, 
while optimizing the duration of testing (\S\ref{sec:model-testing}). 

We have integrated \name with PySyft (\S\ref{sec:implementation}) and evaluated it across various FL tasks with real-world workloads (\S\ref{sec:eval}).
\footnote{\name is available at {\url{https://github.com/SymbioticLab/Oort}}.}
Compared to the state-of-the-art selection techniques used in today's FL deployments \cite{gg-keyeval, ggkeyboard, gg-oov}, \name improves time-to-accuracy performance by 1.2$\times$-14.1$\times$ and final model accuracy by 1.3\%-9.8\% for federated model training, while achieving  close to upper-bound statistical performance. 
For federated model testing, \name can efficiently respond to developer-specified data distribution across millions of clients, and improves the end-to-end testing duration by 4.7$\times$ on average over state-of-the-art solutions.

Overall, we make the following contributions in this paper: 
\begin{denseenum}
  \item We highlight the tension between statistical and systems efficiency when selecting FL participants and present \name to effectively navigate the tradeoff.

  \item We propose participant selection algorithms to improve the time-to-accuracy performance of training and to scalably enforce developers' FL testing criteria.
  
  \item We implement and evaluate these algorithms at scale in \name, showing both statistical and systems performance improvements over the state-of-the-art.

\end{denseenum}

\section{Background and Motivation}
\label{sec:background}
We start with a quick primer on federated learning (\S\ref{sec:bg-fl}), followed by the challenges it faces based on our analysis of real-world datasets (\S\ref{sec:challenges}). 
Next, we highlight the key shortcomings of the state-of-the-art that motivate our work (\S\ref{sec:limitation}).

\subsection{Federated Learning}
\label{sec:bg-fl}

%

Training and testing play crucial roles in the life cycle of an FL model, 
whereas they have different criteria. 

Federated model training aims to learn an accurate model across thousands to potentially millions of clients. 
Because of the large population size and diversity of user data and their devices in FL, training runs on a subset of clients (hundreds of participants) in each round, 
and often takes hundreds of rounds (each round lasts a few minutes) and several days to complete.
For example, in Gboard keyboard, Google runs federated training of NLP models over weeks across 1.5 million end devices \cite{ggkeyboard, gg-nwp}. 
For a given model, achieving a target model accuracy with less wall clock time (\ie, time-to-accuracy) is still the primary target \cite{fl-heternet, fed-yogi}.

To inspect a model's accuracy during training (\eg, to detect cut-off accuracy), to validate the trained model before deployment \cite{gg-keyeval, ggkeyboard, gg-oov}, or to circumvent biased proxy data in hyperparameter tuning \cite{fl-memorize, inspect-fl}, FL developers sometimes test model's performance on real-life datasets. 
Similar to traditional ML, developers often request the representativeness of the testing set with requirements like \textit{``50k representative samples"} \cite{inspect-fl}, or \textit{``x samples of class y"} to investigate model performance on specific categories \cite{positive-fl}. 
When the data characteristics of participants are not available, coarse-grained yet non-trivial requests, such as \textit{``a subset with less than X\% data deviation from the global"} are still informative \cite{model-interpolation, fl-distri}.

\subsection{Challenges in Federated Learning}
\label{sec:challenges}

Apart from the challenges faced in traditional ML, FL introduces new challenges in terms of data, systems, and privacy.

\begin{figure}[t]
  \centering
  {
    \subfigure[Unbalanced data size. \label{fig:data-size}]{\includegraphics[width=0.5\linewidth]{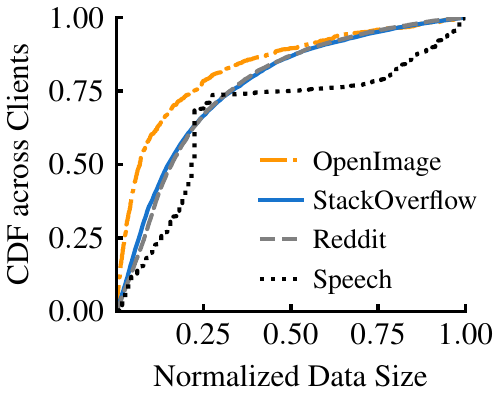}}\hfill
    \subfigure[Heterogeneous data distribution. \label{fig:data-div}]{\includegraphics[width=0.5\linewidth]{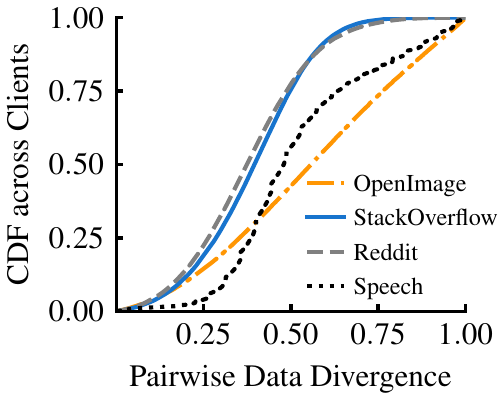}}
  }
  \caption{Client data differs in size and distribution greatly.}
  \label{fig:heter-data}
\end{figure}

\paragraph{Heterogeneous statistical data.}
Data in each FL participant is typically generated in a distributed manner under different contexts and stored independently. 
For example, images collected by cameras will reflect the demographics of each camera’s location.
This breaks down the widely-accepted assumption in traditional ML that samples are independent and identically distributed (i.i.d.) from a data distribution.
  
We analyze four real-world datasets for CV (OpenImage \cite{openimg}) and NLP (StackOverflow \cite{stack-overflow}, Reddit \cite{reddit} and Google Speech \cite{google-speech}) tasks.
Each consists of thousands or up to millions of clients and millions of data points. 
In each individual dataset, we see a high statistical deviation across clients not only in the quantity of samples (Figure~\ref{fig:data-size}) but also in the data distribution (Figure~\ref{fig:data-div}).\footnote{We report the pairwise deviation of categorical distributions between two clients, using the popular L1-divergence metric \cite{probablity}.}

\begin{figure}[t]
  \centering
  {
    \subfigure[Heterogeneous compute capacity. \label{fig:computer-div}]{\includegraphics[width=0.5\linewidth]{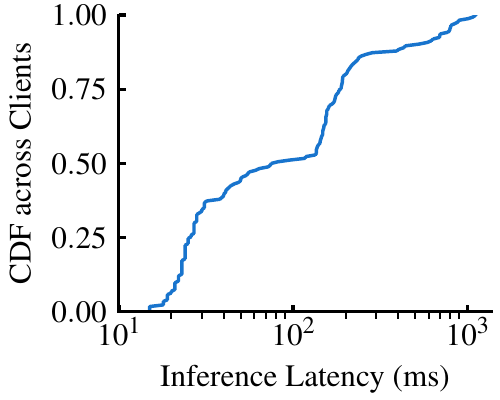}}\hfill
    \subfigure[Heterogeneous network capacity. \label{fig:bw-div}]{\includegraphics[width=0.5\linewidth]{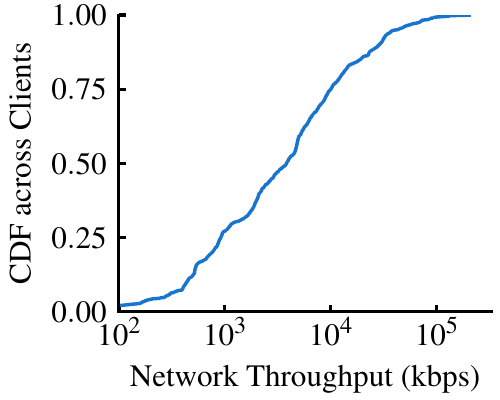}}
  }
  \caption{Client system performance differs significantly.}
  \label{fig:heter-sys}
\end{figure}

\paragraph{Heterogeneous system performance.}
As individual data samples are tightly coupled with the participant device, in-situ computation on this data experiences significant heterogeneity in system performance. 
We analyze the inference latency of MobileNet \cite{mobilenet} across hundreds of mobile phones used in a real-world FL deployment \cite{ggkeyboard}, and their available bandwidth. 
Unlike the homogeneous setting in datacenter ML, system performance across clients exhibits an order-of-magnitude difference in both computational capabilities (Figure~\ref{fig:computer-div}) and network bandwidth (Figure~\ref{fig:bw-div}).

\paragraph{Enormous population and pervasive uncertainty.} 
While traditional ML runs in a well-managed cluster with a number of machines, federated learning often involves up to  millions of clients, making it challenging for the coordinator to efficiently identify and manage valuable participants.
During execution, devices often vary in system performance \cite{fl-survey, federated-learning} -- they may slow down or drop out -- and the model performance varies in FL training as the model updates over rounds.

\paragraph{Privacy concerns.} 
Inquiring about the privacy-sensitive information of clients (\eg, raw data or even data distribution) 
can alienate participants in contributing to FL \cite{apple-dp, gdpr-vldb, gdpr-hotcloud}. 
Hence, realistic FL solutions have to seek efficiency improvements 
but with limited information available in practical FL, 
and their deployments must be non-intrusive to clients. 



\subsection{Limitations of Existing FL Solutions}
\label{sec:limitation}


While existing FL solutions have made considerable progress in tackling some of the above challenges (\S\ref{sec:related}), they mostly rely on hindsight -- given a pool of participants, they optimize model performance \cite{fl-adver, fl-fair} or system efficiency \cite{fedavg} to tackle data and system heterogeneity. 
However, the potential for curbing these disadvantages by cherry-picking participants before execution has largely been overlooked. For example, FL training and testing today still rely on randomly picking participants \cite{federated-learning}, which leaves large room for improvements. 

\begin{figure}[t]
\vspace{.2cm}
  \centering
  \includegraphics[width=\linewidth]{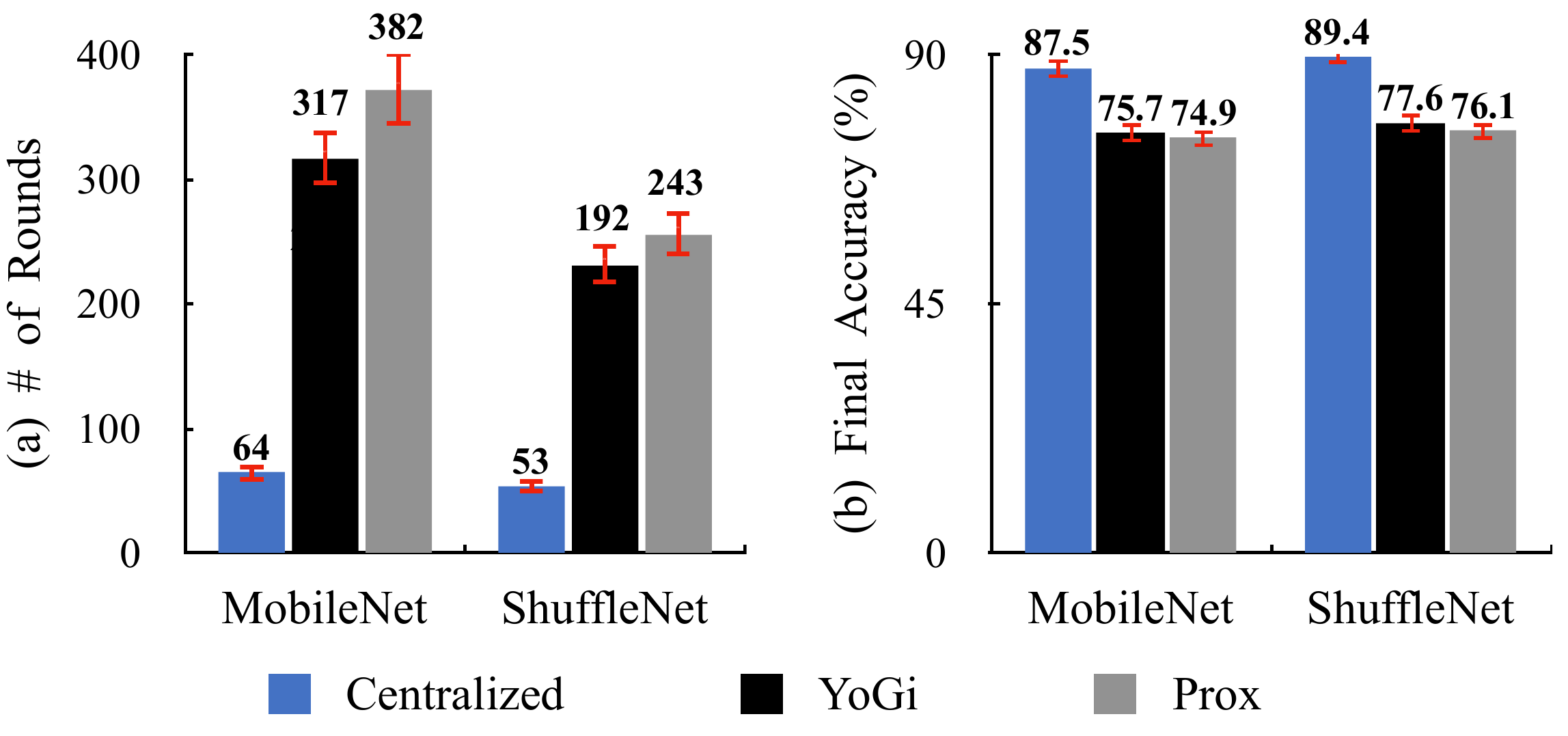}
  \caption{Existing works are suboptimal in: (a) round-to-accuracy performance and (b) final model accuracy. (a) reports number of rounds required to reach the highest accuracy of Prox on MobileNet (\ie, 74.9\%). Error bars show standard deviation.}
  \label{fig:optimal-bar}
\end{figure}

\paragraph{Suboptimality in maximizing efficiency.} 


We first show that today's participant selection underperforms for FL solutions.
Here, we train two popular image classification models tailored for mobile devices (\ie, MobileNet \cite{mobilenet} and ShuffleNet \cite{shufflenet}) with 1.6 million images of the OpenImage dataset, and randomly pick 100 participants out of more than 14k clients in each training round.
We consider a performance \emph{upper bound} by creating a hypothetical centralized case where images are evenly distributed across only 100 clients, and train on all 100 clients in each round.
As shown in Figure~\ref{fig:optimal-bar}, even with state-of-the-art optimizations, such as YoGi \cite{fed-yogi} and Prox \cite{fl-heternet},\footnote{These two adapt traditional stochastic gradient descent algorithms to tackle the heterogeneity of the client datasets.}  the round-to-accuracy and final model accuracy are both far from the upper-bound.
Moreover, overlooking the system heterogeneity can elongate each round, further exacerbating the suboptimality of time-to-accuracy performance.

\paragraph{Inability to enforce data selection criteria.}
While an FL developer often fine-tunes her model by understanding the input dataset, existing solutions do not provide any systems support for her to express and reason about what data her FL model was trained or tested on.
Even worse, existing participant selection not only inflates the execution, 
but can lead to bias and loss of confidence in results \cite{data-val, noniid}.

\begin{figure}[t]
  \centering
  {
    \subfigure[Data deviation vs. participant size. \label{fig:data-sample}]{\includegraphics[width=0.49\linewidth]{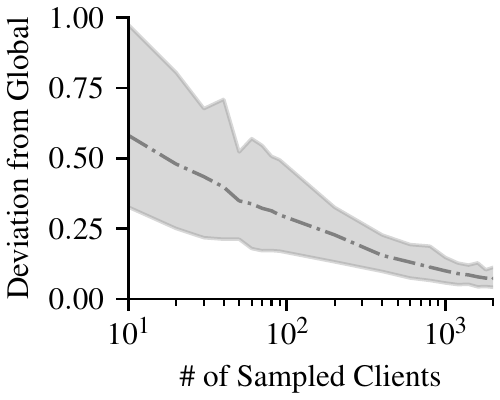}}\hfill
    \subfigure[Accuracy vs. participant size. \label{fig:acc-sample}]{\includegraphics[width=0.49\linewidth]{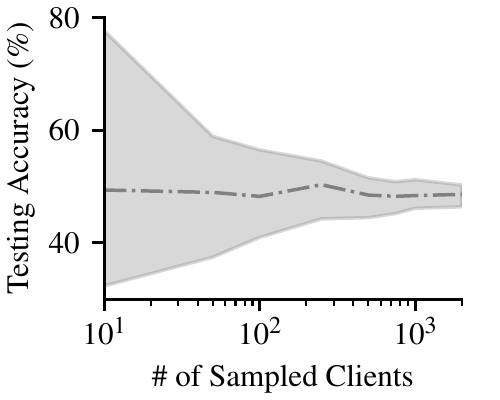}}
  }
  \caption{Participant selection today leads to (a) deviations from developer requirements, and thus (b) affects testing result. 
	Shadow indicates the [min, max] range of y-axis values over 1000 runs given the same x-axis input; each line reports the median.}
  \label{fig:deviation}
\end{figure}

To better understand how existing works fall short, we take the global categorical distribution as an example requirement, and experiment with the above pre-trained ShuffleNet model. 
Figure~\ref{fig:data-sample} shows that: 
(i) even for the same number of participants, random selection can result in noticeable data deviations from the target distribution; 
(ii) while this deviation decreases as more participants are involved, it is non-trivial to quantify how it varies with different number of participants, even if we ignore the cost of enlarging the participant set. 
\update{
Worse, when even selecting many participants, developers can not enforce other distributions (\eg, balanced distribution for debugging \cite{inspect-fl}) with random selection. 
}
One natural effect of violating developer specification is bias in results (Figure~\ref{fig:acc-sample}), 
where we test the accuracy of the same model on these participants. 
We observe that a biased testing set results in high uncertainties in testing accuracy.


\section{\name Overview}
\label{sec:overview}

\name improves FL training and testing performance 
by judiciously selecting participants while enabling FL developers to specify data selection criteria. 
In this section, we provide an overview of how \name fits in the FL life cycle to help the reader follow the subsequent sections.

\subsection{Architecture}
At its core, \name is a participant selection framework that identifies and cherry-picks valuable participants for FL training and testing. 
It is located inside the coordinator of an FL framework and interacts with the driver of an FL execution (\eg, PySyft \cite{pysyft} or Tensorflow Federated \cite{tff}).
Given developer-specified criteria, it responds with a list of participants, whereas the driver is in charge of initiating and managing execution on the \name-selected remote participants. 

\begin{figure}[t]
  \centering
  \includegraphics[width=\linewidth]{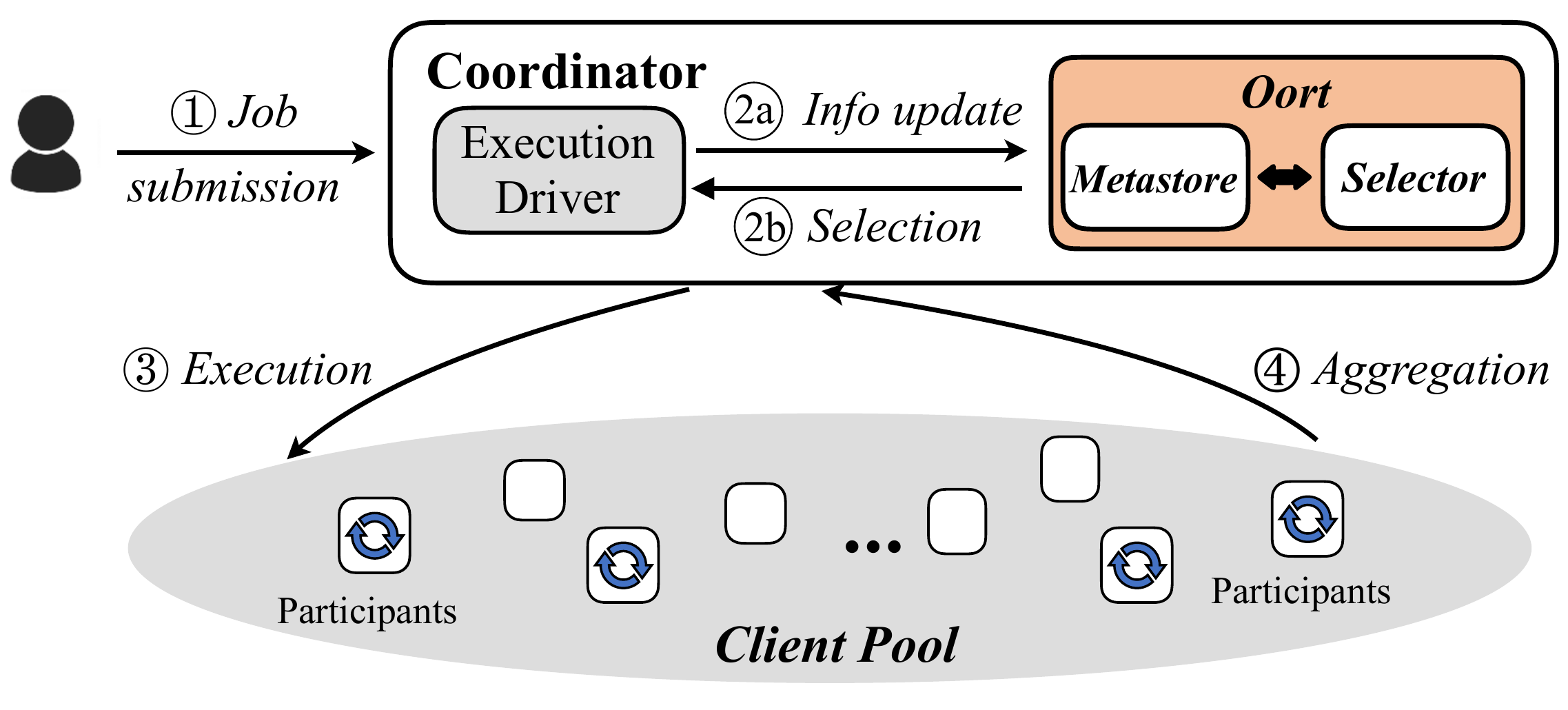}
  \caption{\name architecture. The driver of the FL framework interacts with \name using a client library.}
  \label{fig:overview}
\end{figure}

Figure~\ref{fig:overview} shows how \name interacts with the developer and FL execution frameworks. 
\circled{1} \emph{Job submission}: 
the developer submits and specifies the participant selection criteria to the FL coordinator in the cloud. 
\circled{2} \emph{Participant selection}: 
the coordinator enquires the clients meeting eligibility properties (\eg, battery level), 
and forwards their characteristics (\eg, liveness) to \name. 
Given the developer requirements (and execution feedbacks in case of training \circled{2a}), 
\name selects participants based on the given criteria and notifies the coordinator of this participant selection (\circled{2b}). 
\circled{3} \emph{Execution}: 
the coordinator distributes relevant profiles (\eg, model) to these participants, 
and then each participant independently computes results (\eg, model weights in training) on her data; 
\circled{4} \emph{Aggregation}: 
when participants complete the computation, the coordinator aggregates updates from participants.

During federated training, where the coordinator initiates the next training round after aggregating updates from enough number of participants \cite{federated-learning}, it iterates over \circled{2}-\circled{4} in each round.
Every few training rounds, federated testing is often used to detect whether the cut-off accuracy has been reached. 

\subsection{\name Interface}
\name employs two distinct selectors that developers can access via a client library during FL training and testing.

\paragraph{Training selector.}
This selector aims to improve the time-to-accuracy performance of federated training. 
To this end, it captures the utility of clients in training, and efficiently explores and selects high-utility clients at runtime.

\begin{figure}[h]
\begin{lstlisting}[xleftmargin=3.5ex,language=Python,label={lst:selector},escapechar=|]
import Oort

def federated_model_training():
  selector = Oort.create_training_selector(config)

  # Train to target testing accuracy
  while federated_model_testing() < target:

    # Train 50 rounds before testing
    for _ in range(50):
      # Collect feedbacks of last round
      feedbacks = engine.get_participant_feedback()  |\label{code:training-feedback}| 

      # Update the utility of clients
      for clientId in feedbacks:  |\label{code:training-util}| 
        selector.update_client_util(
                    clientId, feedbacks[clientId]) |\label{code:util-end}| 

      # Pick 100 high-utility participants 
      participants = selector.select_participant(100) |\label{code:training-select}| 
      ... # Activate training on remote clients
\end{lstlisting}
  \caption{Code snippet of \name interaction during FL training.}
  \label{code:training-selector}
\end{figure}

Figure~\ref{code:training-selector} presents an example of how FL developers and frameworks interact with \name during training. 
In each training round, \name collects feedbacks from the engine driver, and updates the utility of individual clients (Line~\ref{code:training-util}-\ref{code:util-end}). 
Thereafter, it cherry-picks high-utility clients to feed the underlying execution (Line~\ref{code:training-select}).
We elaborate more on client utility and the selection mechanism in Section~\ref{sec:model-training}.

\paragraph{Testing selector.}
This selector currently supports two types of selection criteria. 
When the individual client data characteristics (\eg, categorical distribution) are not provided, 
the testing selector determines the number of participants needed to cap the data deviation of participants from the global. 
Otherwise, it cherry-picks participants to serve the exact developer-specified requirements on data while minimizing the duration of testing. 
We elaborate more on selection for federated testing in Section~\ref{sec:model-testing}.


\section{Federated Model Training} 
\label{sec:model-training}


In this section, we first outline the trade-off in selecting participants for FL training 
(\S\ref{sec:design-space}),
and then describe how \name quantifies the client utility while respecting privacy  
(\S\ref{sec:stats-util} and \S\ref{sec:client-util}), how it selects high-utility clients at scale
despite staleness in client utility as training evolves (\S\ref{sec:exp}). 

\subsection{Tradeoff Between Statistical and System Efficiency}
\label{sec:design-space}

Time-to-accuracy performance of FL training relies on two aspects:
(i) \emph{statistical efficiency}: the number of rounds taken to reach target accuracy; and
(ii) \emph{system efficiency}: the duration of each training round.
The data stored on the client and the speed with which it can perform training
determine its utility with respect to statistical and system efficiency, which
we respectively refer to as statistical and system utility.

\begin{figure}[t]
  \centering
  \includegraphics[width=\linewidth]{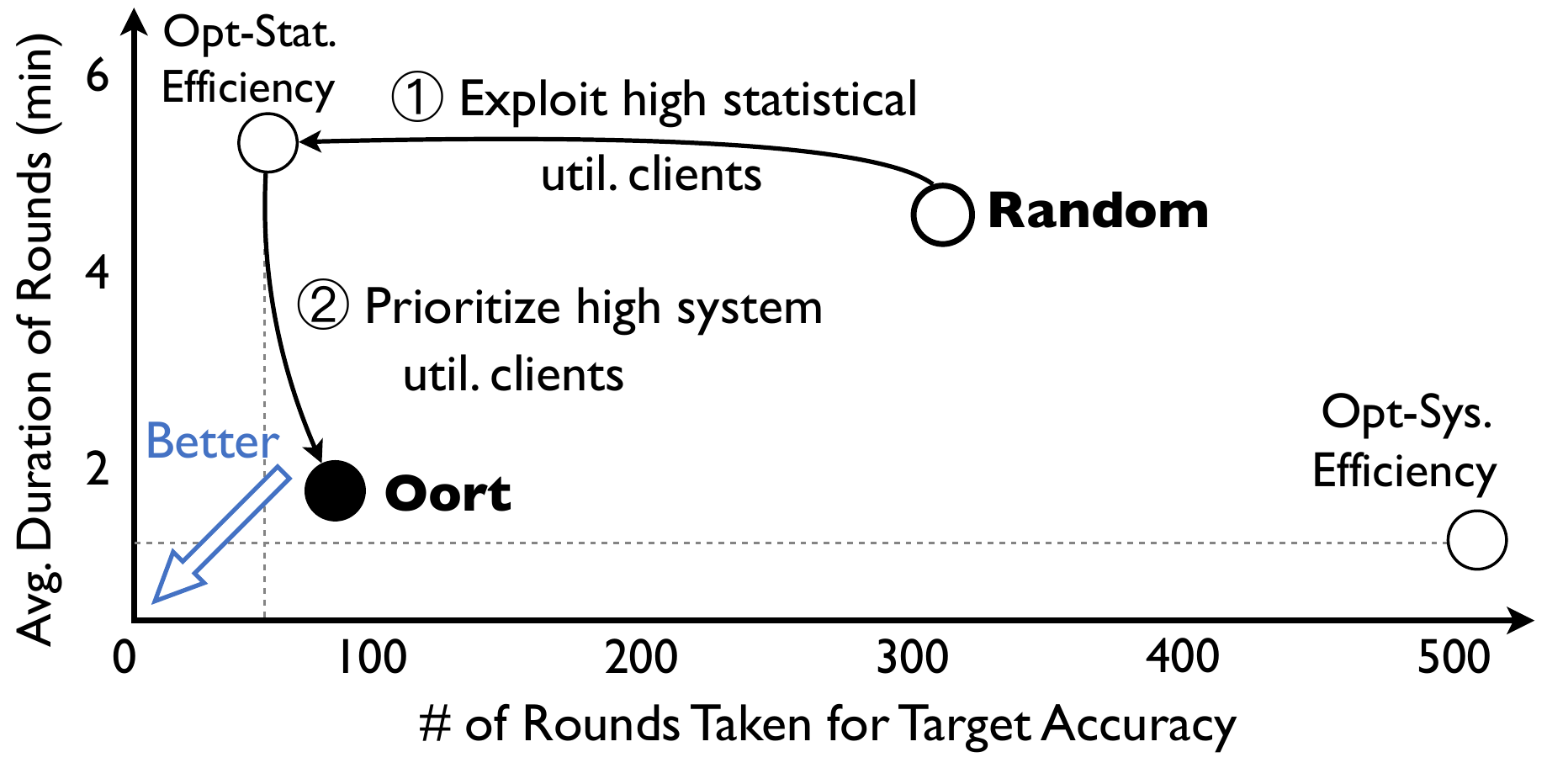}
  \caption{Existing FL training randomly selects participants, whereas \name navigates the sweet point of statistical and 
  system efficiency to optimize their circled area (\ie, time to accuracy). Numbers are from the MobileNet on OpenImage dataset (\S\ref{eval:end-to-end}). }
  \label{fig:trade-off}
\end{figure}

Due to the coupled nature of client data and system performance, 
cherry-picking participants for better time-to-accuracy performance 
requires us to jointly consider both forms of efficiency.   
We visualize the trade-off between these two with our breakdown experiments on the MobileNet model with OpenImage dataset (\S\ref{eval:end-to-end}).  
As shown in Figure~\ref{fig:trade-off}, 
while optimizing the system efficiency (``Opt-Sys. Efficiency'') 
can reduce the duration of each round (\eg, picking the fastest clients), 
it can lead to more rounds than random selection 
as that client data may have already been overrepresented by other participants over past rounds. 
On the other hand, 
using a client with high statistical utility (``Opt-Stat. Efficiency'') may lead to 
longer rounds if that client turns out to be the system bottleneck in global model aggregation. 

\paragraph{Challenges.} 
To improve time-to-accuracy performance, \name aims to find a sweet spot in the trade-off by associating with every client its \emph{utility} 
toward optimizing each form of efficiency (Figure~\ref{fig:trade-off}). 
This leads to three challenges:
\begin{denseitemize}
\item In each round, how to determine which clients' data would help improve 
the statistical efficiency of training the most while respecting client privacy (\S\ref{sec:stats-util})?

\item How to take a client's system performance into account to optimize 
the global system efficiency (\S\ref{sec:client-util})? 

\item How to account for the fact that we don't have up-to-date utility values 
for all clients during training (\S\ref{sec:exp})? 
\end{denseitemize}

Next, we integrate system designs with ML principles to 
tackle the heterogeneity, the massive scale, 
the runtime uncertainties and privacy concerns of clients for practical FL.

\subsection{Client Statistical Utility}
\label{sec:stats-util}

An ideal design of statistical utility should be able to 
efficiently capture the client data utility toward improving model performance 
for various training tasks, and respect privacy.  

To this end, we leverage importance sampling \update{used in the ML literature \cite{is-icml, importance-sampling}}.
Say each client $i$ has a bin $B_i$ of training samples locally stored.
Then, to improve the round-to-accuracy performance via importance sampling, 
the optimal solution would be to pick bin $B_i$ with a
probability proportional to its importance $|B_i| \sqrt{\frac{1}{|B_i|}\sum_{k
\in B_i} \parallel \nabla f(k) \parallel^2}$, where $\parallel \nabla f(k)
\parallel$ is the L2-norm of the unique sample $k$'s gradient $\nabla f(k)$ in
bin $B_i$. Intuitively, this means selecting the bin with larger aggregate
gradient norm across all of its samples.

However, taking this importance as the statistical utility is impractical, since it 
 requires an extra time-consuming pass over the client data to generate the gradient norm of every sample,\footnote{ML models generate the training loss of each sample during training, but calculate the gradient of the mini-batch instead of individual samples.} and this gradient norm varies as the model updates.   

To avoid extra cost, we introduce a pragmatic approximation of statistical utility instead.
At its core, the gradient is derived by taking the derivative of training loss with respect to current model weights, 
wherein training loss measures the estimation error between model predictions and the ground truth.
Our insight is that a larger gradient norm often attributes to a bigger loss \cite{is-nips}.  
Therefore, we 
define the statistical utility $U(i)$ of client $i$ as
$ U(i) = |B_i| \sqrt{\frac{1}{|B_i|}\sum_{k \in B_i} Loss(k) ^2}$,
where the training loss $Loss(k)$ of sample $k$ is automatically generated during training with negligible collection overhead. 
As such, we consider clients that currently accumulate a bigger loss to be more important for future rounds.

Our statistical utility can capture the heterogeneous data utility 
across and within categories and samples for various tasks.  
We present the theoretical insights for its effectiveness over random sampling in Appendix~\ref{app:utility-proof}, 
and empirically show its close-to-optimal performance (\S\ref{eval:eval-breakdown}).

\paragraph{How \name respects privacy?}
\update{
Training loss measures the prediction confidence of a model without revealing the raw data and is often collected in real FL deployments \cite{firefox, ggkeyboard}. 
We further provide three ways to respect privacy. 
First, we rely on {\em aggregate} training loss, which is computed locally by the client across {\em all} of her samples without revealing the loss distribution of individual samples either. 
Second, when even the aggregate loss raises a privacy concern,
clients can add noise to their loss value before uploading, 
similar to existing local differential privacy \cite{diff-fl}.
Third, we later show that \name can flexibly accommodate 
other definitions of statistical utility used in our generic participant selection framework (\S\ref{sec:exp}). 
We provide detailed theoretical analyses for each strategy (\eg, using gradient norm of batches) of 
how \name can respect privacy (\eg, amenable under noisy utility value) in Appendix~\ref{app:privacy}, 
while empirically showing its superior performance even under noisy utility value (\S\ref{eval:sensitivity}).
}

\subsection{Trading off Statistical and System Efficiency}
\label{sec:client-util}

Simply selecting clients with high statistical utility can hamper the system efficiency. 
To reconcile the demand for both efficiencies, we should maximize the statistical utility we can achieve per unit time 
\update{(\ie, the division of statistical utility and its round duration)}. 
As such, we formulate the utility of client $i$ by associating her statistical utility with a 
global system utility in terms of the duration of each training round: 
\begin{align}
 Util(i) = \underbrace{|B_i| \sqrt{\frac{1}{|B_i|}\sum_{k \in B_i} Loss(k) ^2}}_{Statistical \  utility \ U(i)} \times \underbrace{(\frac{T}{t_i})^{\mathbbm{1}(T<t_i) \times \alpha}}_{Global \ sys \ utility}
\label{eq:utility}
\end{align}
where $T$ is the developer-preferred duration of each round, 
$t_i$ is the amount of time that client $i$ takes to process the training,  
which has already been collected by today's coordinator from past
rounds,\footnote{We only care whether a client can complete by the expected
duration $T$. So, a client can even mask its precise speed by deferring its
report.}  
and $\mathbbm{1}(x)$ is an indicator function that takes value 1 if $x$ is true and 0 otherwise. 
This way, the utility of those clients who may be the bottleneck of the desired speed of current round will be penalized by a developer-specified factor $\alpha$, but we do not reward the non-straggler clients because their completions do not impact the round duration. 

This formulation assumes that all samples at a client are processed in that training round.  
Even if the estimated $t_i$ for a client is greater than the desired round 
duration $T$, \name might pick that client if the
statistical utility outweighs its slow speed.  Alternatively, if the
developer wishes to cap every round at a certain duration \cite{fedavg}, 
then either only clients with $t_i <T$ can be considered (\eg, by setting $\alpha \to \infty$) or a subset of a participant's samples can be processed \cite{fl-heternet, fed-yogi}, \update{and only the aggregate training loss of those trained data in that round is considered in measuring the statistical utility.}

\paragraph{Navigating the trade-off.} 
Determining the preferred round duration $T$ in Equation (\ref{eq:utility}),  which strikes the trade-off between the statistical and system efficiency in aggregations, is non-trivial.
Indeed, the total statistical utility (\ie, $\sum{U(i)}$) achieved by picking high utility clients 
 can decrease round by round, 
because the training loss decreases as the model improves over time. 
If we persist in suppressing clients with high statistical utility but low system speed, 
the model may converge to suboptimal accuracy (\S\ref{eval:eval-breakdown}).

To navigate the optimal trade-off -- maximizing the total statistical utility achieved without greatly sacrificing the system efficiency -- \name employs a pacer to determine the preferred duration $T$ at runtime. 
The intuition is that, when the accumulated statistical utility in the past rounds decreases, 
the pacer allows a larger $T \gets T + \Delta $ by $\Delta$ to bargain with the statistical efficiency again.
We elaborate more in Algorithm~\ref{fig:2exp}.

\subsection{Adaptive Participant Selection}
\label{sec:exp}

Given the above definition of client utility, we need to address the following
practical concerns in order to select participants with the highest utility in each 
training round.
\begin{denseitemize}
	\item \emph{Scalability}: a client's utility can only be determined after it has participated in training; how to choose from clients at scale without having to try all clients once?
	\item \emph{Staleness}: since not every client participates in every round, how to account for the change in a client's utility since its last participation?
	\item \emph{Robustness}: how to be robust to outliers in the presence of corrupted clients (\eg, with noisy data)?
\end{denseitemize}

To tackle these challenges, we develop an exploration-exploitation strategy 
for participant selection (Algorithm \ref{fig:2exp}).

\renewcommand*{\algorithmcfname}{Alg.}
\renewcommand{\algcomment}[1]{{\footnotesize\Comment{#1}}}
\begin{algorithm}[!t]
{
  \DontPrintSemicolon
  \SetNoFillComment
  \algrule
  \KwIn{Client set $\mathbb{C}$, sample size $K$, exploitation factor $\epsilon$, pacer step $\Delta$,  step window $W$, penalty $\alpha$}
  \KwOut{Participant set $\mathbb{P}$}
  \algrule
  \SetKwFunction{FMain}{\textrm{SelectParticipant}}
  \SetKwProg{Fn}{Function}{}{}

  \tcc{Initialize global variables.}

  {
	  $\mathbb{E} \gets \emptyset; \ \mathbb{U} \gets \emptyset$	\algcomment{Explored clients and statistical utility.}

	  $\mathbb{L} \gets \emptyset; \ \mathbb{D} \gets \emptyset$	\algcomment{Last involved round and duration.}

	  $R \gets 0; \ T \gets \Delta$	\algcomment{Round counter and preferred round duration.}

  }
  \BlankLine

  \tcc{Participant selection for each round.}
  \Fn{\FMain{$\mathbb{C}$, $K$, $\epsilon$, $T$, $\alpha$}}{
  		{$Util \gets \emptyset; \ R \gets R+1$}

  		\BlankLine
  		\tcc{Update and clip the feedback; blacklist outliers.}
  		UpdateWithFeedback($\mathbb{E}, \ \mathbb{U}, \ \mathbb{L}, \ \mathbb{D}$) \label{alg:update}

  		\BlankLine
  		\tcc{Pacer: Relaxes global system preference $T$ if the statistical utility achieved decreases in last $W$ rounds.}
	    \If{$\sum{\mathbb{U}(R-2W:R-W)}$ > $\sum{\mathbb{U}(R-W:R)}$ \label{alg:pacer-update}} {
	    	$T \gets T + \Delta$
	    }

	    \BlankLine
	    \tcc{Exploitation \#1: Calculate client utility.}	
	    {

			\For{client $i \in \mathbb{E}$ \label{alg:util-tcc}}{
		    	$Util(i) \gets \mathbb{U}(i)$ + $\sqrt{\frac{0.1 \log{R}}{\mathbb{L}(i)}}$ \algcomment{Temporal uncertainty.}	\label{alg:uncertainty}

		    	\If (\algcomment{Global system utility.}) {$T$ < $\mathbb{D}(i)$ \label{alg:sys-pref}}
		    		{$Util(i) \gets Util(i) \times (\frac{T}{\mathbb{D}(i)})^{\alpha}$} \label{alg:cal-util}
		    }
		}

	    \BlankLine
	    \tcc{Exploitation \#2: admit clients with greater than $c$\% of cut-off utility; then sample $(1-\epsilon)K$ clients by utility.}
	    {	
	    	$Util \gets $ SortAsc(Util)	\label{alg:confidence}

		    $\mathbb{W} \gets $ CutOffUtil($\mathbb{E}, \ c \times Util((1-\epsilon)\times K)$)	\label{alg:augment}

		    $\mathbb{P} \gets $ SampleByUtil$(\mathbb{W}, \ Util, \ (1-\epsilon) \times K)$	\label{alg:samplebyutil}
		}

	    \BlankLine
	    \tcc{Exploration: sample unexplored clients by speed.}

	    $\mathbb{P} \gets \mathbb{P}$ $\cup$ SampleBySpeed$(\mathbb{C} - \mathbb{E}, \ \epsilon \times K)$ \label{alg:exploration}

	    \BlankLine
	    \Return $\mathbb{P}$
	}
}
\algrule
\caption{Participant selection w/ exploration-exploitation.}
\label{fig:2exp}
\end{algorithm}

\paragraph{Online exploration-exploitation of high-utility clients.} 
Selecting participants out of numerous clients can be modeled as a multi-armed bandit problem, where each client is an ``arm'' of the bandit, and the utility obtained is the ``reward''~\cite{ml-ucb}. \update{In contrast to sophisticated designs (\eg, reinforcement learning), the bandit model is scalable and flexible even when the solution space (\eg, number of clients) varies dramatically over time. 
Next, we adaptively balance the exploration and exploitation of different arms to maximize the long-term reward.
}


Similar to the bandit design, 
\name efficiently explores potential participants under spatial variation, while intelligently exploiting observed high-utility participants under temporal variation. 
At the beginning of each selection round, \name receives the feedback of the last training round, 
and updates the statistical utility and system performance of clients (Line \ref{alg:update}). 
For the explored clients, \name calculates their client utility and narrows down the selection by exploiting the high-utility participants (Line \ref{alg:util-tcc}-\ref{alg:samplebyutil}).
Meanwhile, \name samples $\epsilon$($\in$ [0, 1]) fraction of participants to explore potential participants that had not been selected before (Line \ref{alg:exploration}), which turns to full exploration as $\epsilon \to 1$. 
Although we cannot learn the statistical utility of not-yet-tried clients, one can decide to prioritize the unexplored clients with faster system speed when possible (\eg, by inferring from device models), instead of performing random exploration (Line \ref{alg:exploration}).


\paragraph{Exploitation under staleness in client utility.} 
\name employs two strategies to account for the dynamics in client utility over time.
First, motivated by the confidence interval used to measure the uncertainty in bandit reward, we introduce an incentive term, which shares the same shape of the confidence in bandit solutions \cite{via}, to account for the staleness (Line \ref{alg:uncertainty}), whereby we gradually increase the utility of a client if she has been overlooked for a long time. So those clients accumulating high utility since their last trial can still be \update{repurposed} again. 
Second, instead of picking clients with top-k utility deterministically, we allow a confidence interval $c$ on the cut-off utility (95\% by default in Line \ref{alg:confidence}-\ref{alg:augment}). 
Namely, we admit clients whose utility is greater than the $c$\% of the top ($(1-\epsilon) \times K$)-th participant.
Among this high-utility pool, \name samples participants with probability proportional to their utility (Line \ref{alg:samplebyutil}).
\update{
This adaptive exploitation mitigates the uncertainties in client utility by prioritizing participants opportunistically, thus relieving the need for accurate estimations of utility as we do not require the exact ordering among clients, while preserving a high quality as a whole.
}

\paragraph{Robust exploitation under outliers.}
Simply prioritizing high utility clients can be vulnerable to outliers in unfavorable settings. 
For example, corrupted clients may have noisy data, leading to high training loss, 
or even report arbitrarily high training loss intentionally.
For robustness, \name 
(i) removes the client in selection after she has been picked over a given number of rounds.
This helps to remove the perceived outliers in terms of participation (Line \ref{alg:update});
(ii) clips the utility value of a client by capping it to no more than an upper bound (\eg, 95\% value in utility distributions). With probabilistic participant selection among the high-utility client pool (Line \ref{alg:samplebyutil}), the chance of selecting outliers is significantly decreased under the scale of clients in FL.
We show that \name outperforms existing mechanisms while being robust (\S\ref{eval:sensitivity}).

\paragraph{Accommodation to diverse selection criteria.}
Our adaptive participant selection is generic for different utility definitions of diverse selection criteria. 
For example, 
developers may hope to reconcile their demand for time-to-accuracy efficiency and fairness, 
so that some clients are not underrepresented (\eg, more fair resource usage across clients) \cite{fl-survey, fl-fair}. 
Although developers may have various fairness criterion $fairness(\cdot)$, 
\name can enforce their demands by replacing the current utility definition of client $i$ with $(1-f)\times Util(i) + f \times fairness(i)$, where $f \in [0, 1]$ and Algorithm~\ref{fig:2exp} will naturally prioritize clients with the largest fairness demand as $f \to 1$. 
For example, $fairness(i)= max\_resource\_usage - {resource\_usage(i)}$ motivates fair resource usage for each client $i$. Note that existing participant selection provides no support for fairness, and we show that \name can efficiently enforce diverse developer-preferred fairness while improving performance (\S\ref{eval:sensitivity}).

\section{Federated Model Testing}
\label{sec:model-testing} 

\begin{figure}[t]
\begin{lstlisting}[xleftmargin=3.5ex,language=Python,label={lst:selector},escapechar=|]
def federated_model_testing():
  selector = Oort.create_testing_selector()

  # Type 1: subset w/ < X deviation from the global
  participants = selector.select_by_deviation(
    dev_target, range_of_capacity, total_num_clients)

  # Provide individual client data characteristics
  selector.update_client_info(client_id, client_info)
  # Type 2: [5k, 5k] samples of category [i, j]
  participants = selector.select_by_category(
                 request_list, testing_config)
\end{lstlisting}
  \caption{Key \name APIs for supporting federated testing.}
  \label{code:testing-apis} 
\end{figure}

Enforcing developer-defined requirements on data distribution 
is a first-order goal in FL testing, 
whereas existing mechanisms lead to biased testing results (\S\ref{sec:limitation}).
In this section, we elaborate on how \name serves the two primary types of queries. 
As shown in Figure~\ref{code:testing-apis}, we start with how \name preserves the representativeness of testing set 
even without individual client data characteristics (\S\ref{sec:non-clairvoyant}), 
and how it efficiently enforces developer's testing criteria for specific data distribution when the individual information is provided (\S\ref{sec:clairvoyant}).

\subsection{Preserving Data Representativeness}
\label{sec:non-clairvoyant}

Learning the individual data characteristics (\eg, categorical distribution) 
can be too expensive or even prohibited \cite{honeycrisp, rappor}.
Without knowing data characteristics, 
the developer has to be conservative and selects many participants to gain more 
confidence for query \textit{``a testing set with less than X\% data deviation from the global"}, 
as selecting too few can lead to a biased testing result (\S\ref{sec:limitation}).
However, admitting too many may inflate the budget and/or take too long because of the system heterogeneity. 
Next, we show how \name can enable guided participant selection by determining the number of participants needed to guarantee this deviation target.

We consider the deviation of the data formed by all participants from the global dataset (\ie, representative) using  L1-distance, a popular distance metric in FL \cite{noniid, fl-distri, fedvc}.
For category $X$, its L1-distance ($|\bar{X}-{E}[\bar{X}]|$) captures how the average number of samples of all participants (\ie, empirical value $\bar{X}$) deviates from that of all clients (\ie, expectation ${E}[\bar{X}]$). 
\update{Note that the number of samples $X_{n}$ that client $n$ holds is independent across clients. Namely, the number of samples that one client holds will not be affected by the selection of any other clients at that time, so it can be viewed as a random instance sampled from the distribution of variable $X$. }

Given the developer-specified tolerance $\epsilon$ on data deviation and confidence interval $\delta$ (95\% by default \cite{probability}), our goal is to estimate the number of participants needed such that the deviation from the representative categorical distribution is bounded (\ie, $Pr[|\bar{X}-{E}[\bar{X}]| < \epsilon] > \delta$). 
%
To this end, we formulate it as a problem of sampling stochastic variables, 
\update{
and apply the Hoeffding bound \cite{concentration} to capture how this data deviation varies with different number of participants. 
We attach our theoretical results and proof in Appendix~\cite{app:group-size}.
}

\paragraph{Estimating the number of participants to cap deviation.} 

Even when the individual data characteristics are not available, 
the developer can specify her tolerance $\epsilon$ on the deviation from the global categorical distribution, 
whereby \name outputs the number of participants needed to preserve this preference. 
To use our model, the developer needs to input the global range (\ie, global maximum - global minimum) of the number of samples that one client can hold, and the total number of clients. 
Learning this global information securely is well-established \cite{prio, honeycrisp}, 
and the developer can assume a plausible limit (\eg, according to the capacity of device models) too.

Our model does not require any collection of the distribution of global or participant data. 
As a straw-man participant selection design, 
the developer can randomly distribute her model to this \name-determined number of participants.  
After collecting results from this number of participants, 
she can confirm the representativeness of computed data. 

\subsection{Enforcing Diverse Data Distribution}
\label{sec:clairvoyant}

When the individual data characteristics are provided (\eg, FL across enterprise AI cameras~\cite{fed-oj, fedvc}), 
\name can enforce the exact data preference on specific categorical distribution,  
and improve the duration of testing by cherry-picking participants.

Satisfying queries like \textit{``[5k, 5k] samples of class [x, y]"} can be viewed as a multi-dimensional bin covering problem, where a subset of data bins (\ie, participants) are selected to cover the requested quantity of data. 
\update{
For each category $i(\in I)$ of interest, the developer has preference $p_i$ (preference constraint), and an upper limit $B$ (referred to as budget) on how many participants she can have \cite{inspect-fl}.
%
Each participant $n(\in {N})$ can contribute $n_i$ samples out of her capacity $c{_n}{^i}$ (capacity constraint). 
Given her compute speed $s_n$, the available bandwidth $b_n$ and the size of data transfers $d_n$, 
we aim to minimize the duration of model testing:}
%
\begin{align}
&\min\Big{\{\max\limits_{n \in N}\Big({\frac{\sum_{i \in I}{n_i}}{s_n} + \frac{d_n}{b_n}}\Big)}\Big\} \tag*{$\triangleright$ Minimize duration} \\[-.6\jot] 
\textrm{s.t.} \quad &\forall i \in I, \sum_{n \in N} n_i = p_i \tag*{$\triangleright$ Preference Constraint}\\[-.7\jot] 
&\forall i \in I, \forall n \in N, n_i \leq c_n^i \tag*{$\triangleright$ Capacity Constraint}\\[-.4\jot] 
&\forall i \in I, \sum_{n \in N} \mathbbm{1}(n_i > 0) \leq B \tag*{$\triangleright$ Budget Constraint} 
\label{eq:lp-testing}
\end{align} 
%

\update{The max-min formulation stems from the fact that testing completes after aggregating results from the last participant.
While this mixed-integer linear programming (MILP) model provides high-quality solutions, it has prohibitively
high computational complexity for large $N$.}

\paragraph{Scalable participant selection.} 
For better scalability, we present a greedy heuristic to scale down the search space of this strawman.
We  
(1) first group a subset of feasible clients to satisfy the preference constraint. 
To this end, we iteratively add to our subset the client 
which has the most number of samples across all not-yet-satisfied categories, 
and deduct the preference constraint on each category by the corresponding capacity of this client. 
We stop this greedy grouping until the preference is met, or request a new budget if we exceed the budget;  
and (2) then optimize job duration with a simplified MILP among this subset of clients, 
%
wherein we have removed the budget constraint and reduced the search space of clients.
We show that our heuristic can outperform the straw-man MILP model in terms of the end-to-end duration of model testing owing to its small overhead (\S\ref{eval:testing-clair}).








\section{Implementation}
\label{sec:implementation}

We have implemented \name as a Python library, with 2617 lines of code, to friendly support FL developers. 
\name provides simple APIs to abstract away the problem of participant selection, 
and developers can import \name in their application codebase and interact with FL engines (\eg, PySyft \cite{pysyft} or TensorFlow Federated \cite{tff}). 

We have integrated \name with PySyft. \name operates on and updates its client metadata (\eg, data distribution or system performance) fed by the FL developer and PySyft at runtime. 
The metadata of each client in \name is an object with a small memory footprint. 
\name caches these objects in memory during executions and periodically backs them up to persistent storage. 
In case of failures, the execution driver will initiate a new \name selector, 
and load the latest checkpoint to catch up. 
We employ Gurobi solver \cite{gurobi} to solve the MILP. 
The developer can also initiate a \name application beyond coordinators to avoid resource contention. 
We use \textit{xmlrpc} library to connect to the coordinator,  
and these updates will activate \name to write these updates to its metastore. 
In the coordinator, we use the PySyft API \textit{model.send(client\_id)} to direct which client to run given the \name decision, and \textit{model.get(client\_id)} to collect the feedback. 

\section{Evaluation}
\label{sec:eval}

We evaluate \name's effectiveness for four different ML models on four CV and NLP datasets. 
We organize our evaluation by the FL activities with the following key results.

\paragraph{FL training results summary:}
\begin{denseitemize}
  \item \name outperforms existing random participant selection by 1.2$\times$-14.1$\times$ in time-to-accuracy performance, while achieving 1.3\%-9.8\% better final model accuracy (\S\ref{eval:end-to-end}). 

  \item \name achieves close-to-optimal model efficiency by adaptively striking the trade-off between statistical and system efficiency with different components (\S\ref{eval:eval-breakdown}).
  
  \item \name outperforms its counterpart over a wide range of parameters and  
  different scales of experiments, while being robust to outliers (\S\ref{eval:sensitivity}).

\end{denseitemize}

\paragraph{FL testing results summary:}
\begin{denseitemize}
  \item \name can serve testing criteria on data deviation
  while reducing costs by bounding the number of participants needed without individual data characteristics (\S\ref{eval:non-clair}). 

  \item With the individual information, \name improves the testing duration by 4.7$\times$ w.r.t. Mixed Integer Linear Programming (MILP) solver, and is able to efficiently enforce developer preferences across millions of clients (\S\ref{eval:testing-clair}). 
  
\end{denseitemize}


\subsection{Methodology}
\label{eval:setup}

\paragraph{Experimental setup.}
\name is designed to operate in large deployments with potentially millions of edge devices.
However, such a deployment is not only prohibitively expensive, 
but also impractical to ensure the reproducibility of experiments. 
As such, we resort to a cluster with 68 NVIDIA Tesla P100 GPUs, 
and emulate up to 1300 participants in each round. 
\update{
We simulate real-world heterogeneous client system performance and data in both training and testing evaluations using an open-source FL benchmark~\cite{fedscale}: (1) Heterogeneous device runtimes of different models, network throughput/connectivity, device model and availability are emulated using data from AI Benchmark \cite{ai-bench} and Network Measurements on mobiles \cite{mobiperf}; (2) We distribute each real dataset to clients following the corresponding raw placement (\eg, using \emph{<authors\_ID>} to allocate OpenImage), where client data can vary in quantities, distribution of outputs and input features; (3) The coordinator communicates with clients using the parameter server architecture. These follow the PySyft and real FL deployments. 
To mitigate stragglers, 
we employ the widely-used mechanism specified in real FL deployments \cite{federated-learning}, 
where we collect updates from the first $K$ completed participants out of 1.3$K$ participants in each round, and $K$ is 100 by default. 
We report the simulated clock time of clients in evaluations. 
}

\begin{table}[t]
    \centering 
    \setlength\arrayrulewidth{1.2pt}
    \renewcommand{\arraystretch}{1.}
    \setlength{\tabcolsep}{10pt} 
    \begin{tabularx}{\linewidth}{c c c} 
    \hline
    Dataset & \# of Clients & \# of Samples\\ [.8ex] 
    \hline 

    Google Speech \cite{google-speech} & 2,618 & 105,829\\ [.8ex]
    OpenImage-Easy \cite{openimg} & 14,477 & 871,368\\ [.8ex]
    OpenImage \cite{openimg} & 14,477 & 1,672,231\\ [.8ex]
    StackOverflow \cite{stack-overflow} & 315,902 &  135,818,730\\ [.8ex]
    Reddit \cite{reddit} & 1,660,820 &  351,523,459 \\
    \hline 
    \end{tabularx}
    \caption{Statistics of the dataset in evaluations.} 
    \label{table:data-stats} 
\end{table}

\renewcommand{\arraystretch}{1.58}
\definecolor{Gray}{gray}{0.9}

\begin{table*}[t]
\centering 
\small
\setlength{\arrayrulewidth}{0.9pt}
\begin{tabularx}{\linewidth}{c c c c c c c c c c} 
\hhline{-|-|-|-|-|-|-|-|-|-|} 
\multirow{2}{*}{Task} & \multirow{2}{*}{Dataset} & Accuracy & \multirow{2}{*}{Model} & \multicolumn{3}{c}{Speedup for Prox \cite{fl-heternet}} & \multicolumn{3}{c}{Speedup for YoGi \cite{fed-yogi}} \\ [.6ex] \cline{5-10}  
  & & Target & & Stats. & Sys. & Overall & Stats. & Sys.  & Overall \\
\hhline{-|-|-|-|-|-|-|-|-|-|} 
  & \multirow{2}{*}{OpenImage-Easy \cite{openimg}} &\multirow{2}{*}{74.9\%}  &MobileNet \cite{mobilenet} & 3.8$\times$ & 3.2$\times$  & \cellcolor{Gray} 12.1$\times$ & 2.4$\times$ & 2.4$\times$ & \cellcolor{Gray} 5.7$\times$ \\ \hhline{~~~|-|-|-|-|-|-|-|}
Image &                                   &  &   ShuffleNet \cite{shufflenet}  & 2.5$\times$ & 3.5$\times$ & \cellcolor{Gray} 8.8$\times$ & 1.9$\times$ & 2.7$\times$ & \cellcolor{Gray} 5.1$\times$ \\
  \hhline{~|-|-|-|-|-|-|-|-|-|}
Classification  & \multirow{2}{*}{OpenImage \cite{openimg}} &\multirow{2}{*}{53.1\%}  &MobileNet  & 4.2$\times$ & 3.1$\times$  & \cellcolor{Gray} 13.0$\times$ & 2.3$\times$ & 1.5$\times$ & \cellcolor{Gray} 3.3$\times$ \\ \hhline{~~~|-|-|-|-|-|-|-|}
 &                                   &  &   ShuffleNet    & 4.8$\times$ & 2.9$\times$ & \cellcolor{Gray} 14.1$\times$ & 1.8$\times$ & 3.2$\times$ & \cellcolor{Gray} 5.8$\times$ \\
\hhline{-|-|-|-|-|-|-|-|-|-|}
\multirow{2}{*}{Language Modeling} &  Reddit \cite{reddit} & 39 perplexity & Albert \cite{albert} & 1.3$\times$ & 6.4$\times$ & \cellcolor{Gray} 8.4$\times$ & 1.5$\times$ & 4.9$\times$ & \cellcolor{Gray} 7.3$\times$ \\
\hhline{~|-|-|-|-|-|-|-|-|-|}
                  & StackOverflow \cite{stack-overflow} &  39 perplexity  &  Albert   & 
2.1$\times$ & 4.3$\times$ & \cellcolor{Gray} 9.1$\times$ & 1.8$\times$ & 4.4$\times$ & \cellcolor{Gray} 7.8$\times$ \\
\hhline{-|-|-|-|-|-|-|-|-|-|}  
Speech Recognition  & Google Speech \cite{google-speech}   & 62.2\%  & ResNet-34  \cite{resnet} & 1.1$\times$ & 1.1$\times$ & \cellcolor{Gray} 1.2$\times$ & 1.2$\times$ & 1.1$\times$ & \cellcolor{Gray} 1.3$\times$ \\
\hhline{-|-|-|-|-|-|-|-|-|-|} 
\end{tabularx}
\caption{Summary of improvements on time to accuracy.\protect\footnotemark 
We tease apart the overall improvement with statistical and system ones, and take the highest accuracy that Prox can achieve as the target, which is moderate due to the high task complexity and lightweight models.} 
\label{table:e2e-perf}
\end{table*}

\paragraph{Datasets and models.}
We run three categories of applications with four real-world datasets of different scales, \update{and Table~\ref{table:data-stats} reports the statistics of each dataset}:
\begin{denseitemize}
  \item \emph{Speech Recognition}: the small-scale Google speech dataset \cite{google-speech}. We train a convolutional neural network model (ResNet-34 \cite{resnet}) to recognize the command among 35 categories. 

  \item \emph{Image Classification}: the middle-scale OpenImage \cite{openimg} dataset, with 1.5 million images spanning 600 categories, and a simpler dataset (OpenImage-Easy) with images from the most popular 60 categories. We train MobileNet \cite{mobilenet} and ShuffleNet \cite{shufflenet} models to classify the image.
  
  \item \emph{Language Modeling}: the large-scale StackOverflow\cite{stack-overflow} and Reddit\cite{reddit} dataset. We train next word predictions with Albert model \cite{albert} using the top-10k popular words.
  
\end{denseitemize}  

These applications are widely used in real end-device applications \cite{mobile-mlapps}, and these models are designed to be lightweight.

\paragraph{Parameters.}
The minibatch size of each participant is 16 in speech recognition, and 32 in other tasks. 
The initial learning rate for Albert model is 4e-5, and 0.04 for other models. 
These configurations are consistent with those reported in the literature \cite{gg-gboard}. 
In configuring the training selector, \name uses the popular time-based exploration factor \cite{ml-ucb}, 
where the initial exploration factor is 0.9, and decreased by a factor 0.98 after each round when it is larger than 0.2. 
The step window of pacer $W$ is 20 rounds. We set the pacer step $\Delta$ in a way that it can cover the duration of next $W \times K$ clients in the descending order of explored clients' duration, and the straggler penalty $\alpha$ to 2. 
We remove a client from \name's exploitation list once she has been selected over 10 times.

\paragraph{Metrics.}
We care about the \emph{time-to-accuracy} performance and \emph{final model accuracy} of model training tasks \update{on the testing set}. 
For model testing, we measure the \emph{end-to-end} testing duration, 
which consists of the computation overhead of the solution and the duration of actual computation.

For each experiment, we report the mean value over 5 runs, and error bars show the standard deviation.

\subsection{FL Training Evaluation}
\label{eval:training}

In this section, we evaluate \name's performance on model training, 
and employ Prox \cite{fl-heternet} and YoGi \cite{fed-yogi}.
We refer Prox as Prox running with existing random participant selection, and Prox + \name is Prox running atop \name. 
We use a similar denotation for YoGi.
Note that Prox and YoGi optimize the statistical model efficiency for the given participants, 
while \name cherry-picks participants to feed them.

\footnotetext{\update{We set the target accuracy to be the highest achievable accuracy by all used strategies, which turns out to be Prox accuracy. Otherwise, some may never reach that target. }}

\subsubsection{End-to-End Performance}
\label{eval:end-to-end}

\begin{figure}[!t]
  \centering
  {
    \subfigure[MobileNet (Image). \label{fig:e2e-shufflenet}]{\includegraphics[width=0.49\linewidth]{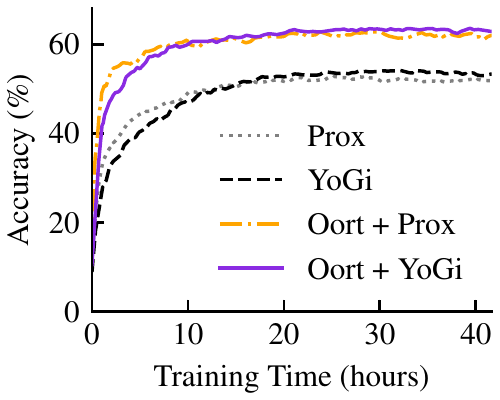}}
    \subfigure[ShuffleNet (Image). \label{fig:e2e-shufflenet}]{\includegraphics[width=0.49\linewidth]{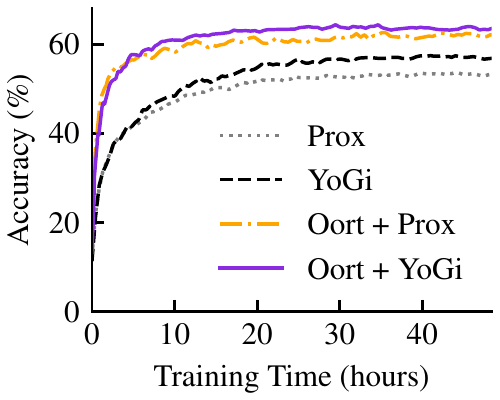}}
    \\
    \subfigure[ResNet (Speech). \label{fig:e2e-resnet}]{\includegraphics[width=0.49\linewidth]{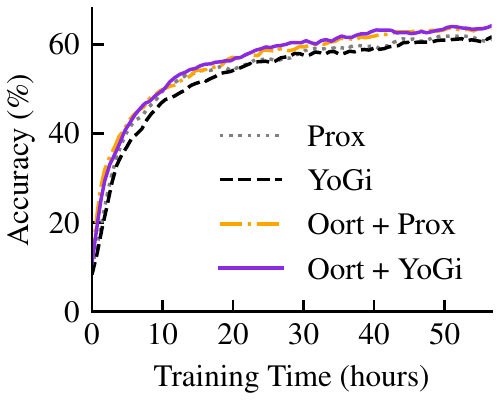}}\hfill
    \subfigure[Albert (LM). \label{fig:e2e-albert}]{\includegraphics[width=0.49\linewidth]{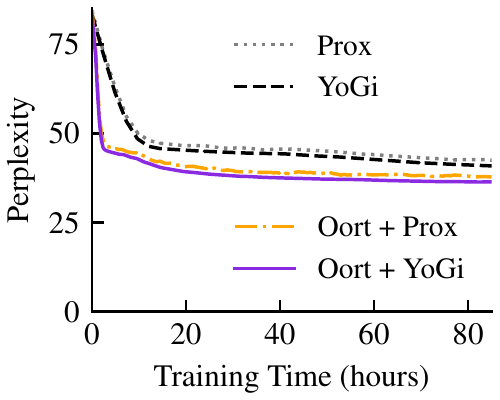}}
  }
  \caption{Time-to-Accuracy performance. A lower perplexity is better in the language modeling (LM) task.}
  \label{fig:end-to-end}
\end{figure}

Table~\ref{table:e2e-perf} summarizes the key time-to-accuracy performance of all datasets. 
In the rest of the evaluations, we report the ShuffleNet and MobileNet performance on OpenImage, and Albert performance on Reddit dataset for brevity. Figure~\ref{fig:end-to-end} reports the timeline of training to achieve different accuracy.  

\begin{figure*}[t]
  \centering
    \subfigure[MobileNet (Image). \label{fig:break-mobilenet}]{\includegraphics[width=0.32\linewidth]{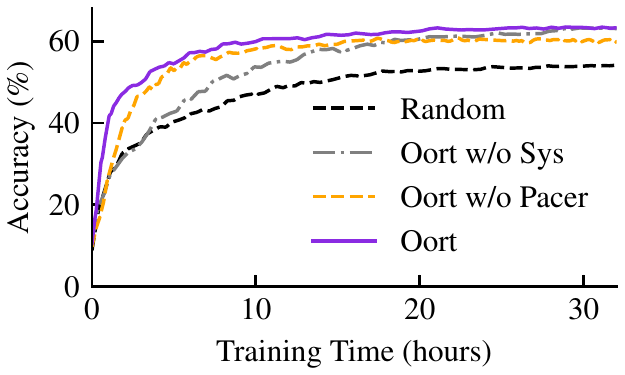}}\hfil
    \subfigure[ShuffleNet (Image). \label{fig:break-shufflenet}]{\includegraphics[width=0.32\linewidth]{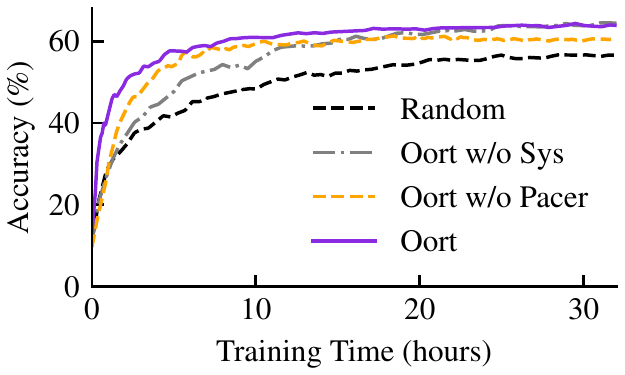}}\hfil 
    \subfigure[Albert (LM). \label{fig:break-albert}]{\includegraphics[width=0.32\linewidth]{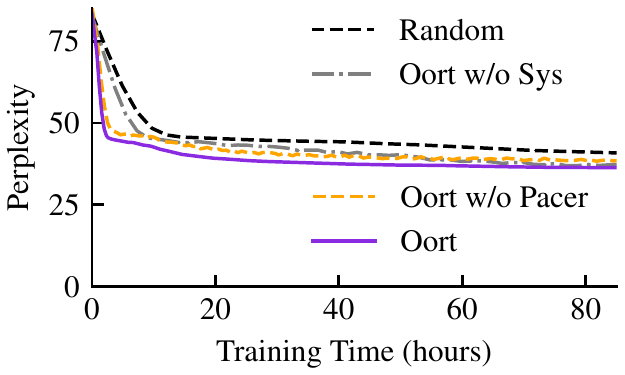}}
  \caption{Breakdown of Time-to-Accuracy performance with YoGi, when using different participant selection strategies.}
  \label{fig:breakdown-tta}
\end{figure*}

\paragraph{\name improves time-to-accuracy performance.}
We notice that \name achieves large speedups to reach the target accuracy (Table~\ref{table:e2e-perf}).
\name reaches the target 3.3$\times$-14.1$\times$ faster in terms of wall clock time on the middle-scale OpenImage dataset; speedup on the large-scale Reddit and StackOverflow dataset is 7.3$\times$-9.1$\times$. 
Understandably, these benefits decrease when the total number of clients is small, 
as shown on the small-scale Google Speech dataset (1.2$\times$-1.3$\times$).

These time-to-accuracy improvements stem from the comparable benefits in statistical model efficiency and system efficiency (Table \ref{table:e2e-perf}).
\name takes 1.8$\times$-4.8$\times$ fewer training rounds on OpenImage dataset to reach the target accuracy, which is better than that of language modeling tasks (1.3$\times$-2.1$\times$).
This is because real-life images often exhibit greater heterogeneity in data characteristics than the language dataset, 
whereas the large population of language datasets leaves a great potential to prioritize clients with faster system speed.

\paragraph{\name improves final model accuracy.}
When the model converges, \name achieves 6.6\%-9.8\% higher final accuracy on OpenImage dataset, 
and 3.1\%-4.4\% better perplexity on Reddit dataset (Figure~\ref{fig:end-to-end}). 
Again, this improvement on Google Speech dataset is smaller (1.3\% for Prox and 2.2\% for YoGi) due to the small scale of clients.
These improvements attribute to the exploitation of high statistical utility clients. 
Specifically, 
the statistical model accuracy is determined by the quality of global aggregation. 
Without cherry-picking participants in each round, 
clients with poor statistical model utility can dilute the quality of aggregation.
As such, the model may converge to suboptimal performance. 
Instead, models running with \name concentrate more on clients with high statistical utility, 
thus achieving better final accuracy.

\subsubsection{Performance Breakdown}
\label{eval:eval-breakdown}

We next delve into the improvement on middle- and large-scale datasets, 
as they are closer to real FL deployments. 
We break down our knobs designed for striking the balance 
between statistical and system efficiency: 
(i) (\emph{\name\ w/o Pacer}): We disable the pacer that guides the aggregation efficiency. 
As such, it keeps suppressing low-speed clients, 
and the training can be restrained among low-utility but high-speed clients;
(ii) (\emph{\name\ w/o Sys}): We further totally remove our benefits from system efficiency by setting $\alpha$ to 0, so \name blindly prioritizes clients with high statistical utility.
We take YoGi for analysis, because it outperforms Prox most of the time.

\paragraph{Breakdown of time-to-accuracy efficiency.}
Figure~\ref{fig:breakdown-tta} reports the breakdown of time-to-accuracy performance, where \name achieves comparable improvement from statistical and system optimizations. 
Taking Figure~\ref{fig:break-shufflenet} as an example,   
(i) At the beginning of training, both \name and (\name\ w/o Pacer) improve the model accuracy quickly, 
because they penalize the utility of stragglers and select clients with higher statistical utility and system efficiency. 
In contrast, (\name\ w/o Sys) only considers the statistical utility, resulting in longer rounds.  
(ii) As training evolves, the pacer in \name gradually relaxes the constraints on system efficiency, 
and admits clients with relatively low speed but higher statistical utility, which ends up with the similar final accuracy of (\name\ w/o Sys). 
However, (\name\ w/o Pacer) relies on a fixed system constraint and suppresses valuable clients with high statistical utility but low speed, leading to suboptimal final accuracy.

\begin{figure}[!t]
\vspace{0.3cm}
  \centering
  \includegraphics[width=0.98\linewidth]{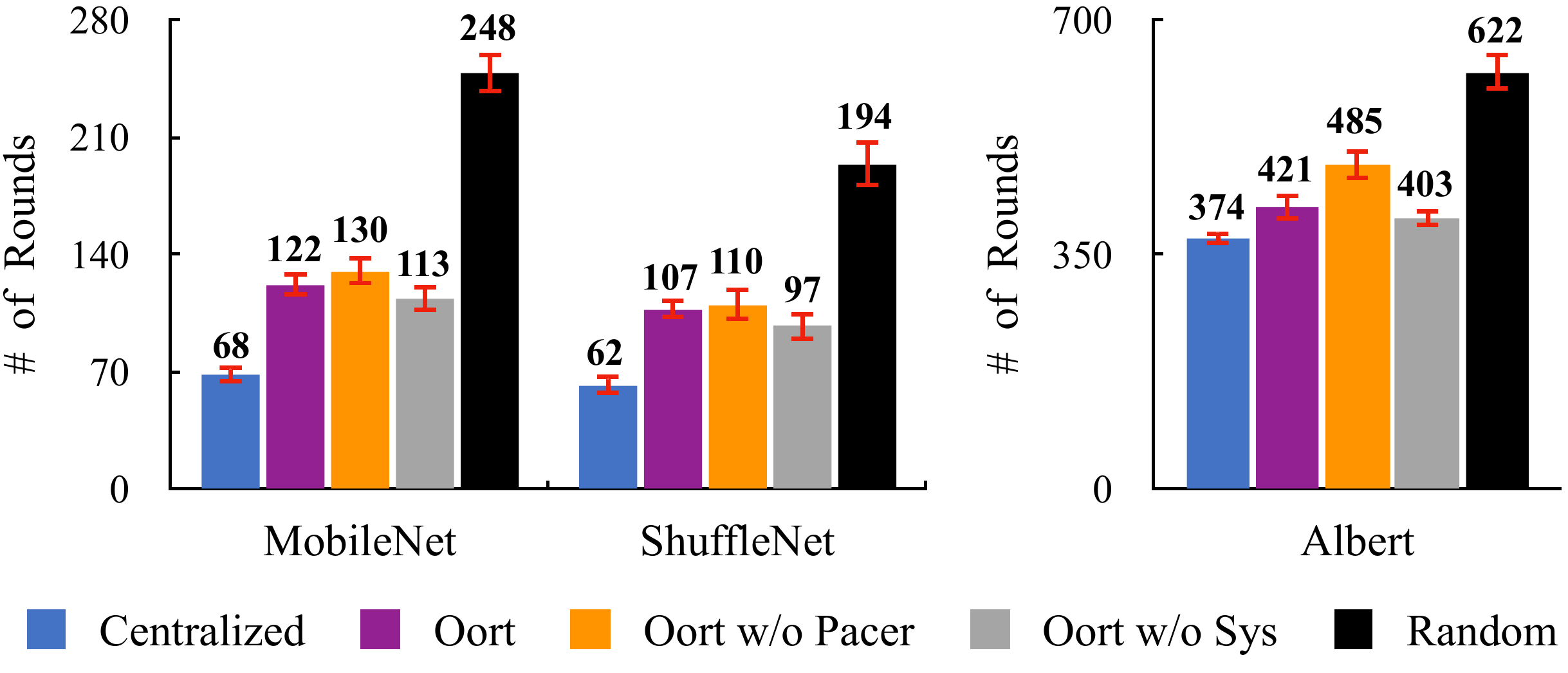}
  \caption{Number of rounds to reach the target accuracy.}
  \label{fig:breakdown-round-efficiency}
\end{figure}

\begin{figure}[!t]
  \centering
  \includegraphics[width=0.98\linewidth]{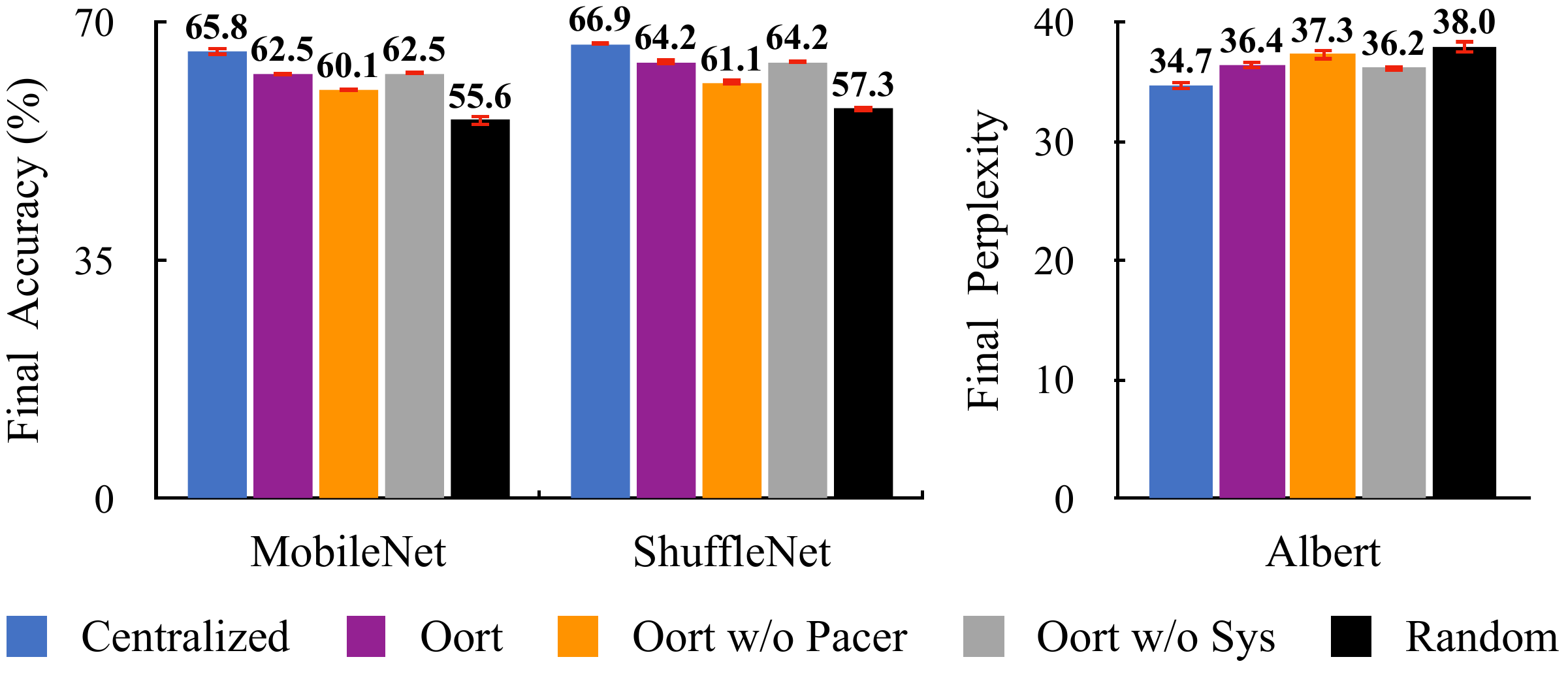}
  \caption{Breakdown of final model accuracy.}
  \label{fig:breakdown-final-accuracy}
\end{figure}

\paragraph{\update{\name achieves close to upper-bound statistical performance.}} 
We consider an \emph{upper-bound} statistical efficiency by creating a centralized case, 
where all data are evenly distributed to $K$ participants. 
Using the target accuracy in Table \ref{table:e2e-perf}, \name can efficiently approach this upper bound by incorporating different components (Figure~\ref{fig:breakdown-round-efficiency}). 
\name is within 2$\times$ of the upper-bound to achieve the target accuracy, and (\name w/o Sys) performs the best in statistical model efficiency, 
because (\name w/o Sys) always grasps clients with higher statistical utility.   
However, it is suboptimal in our targeted time-to-accuracy performance because of ignoring the system efficiency. 
Moreover, by introducing the pacer, \name achieves 2.4\%-3.1\% better accuracy than (\name w/o Pacer), and is merely about 2.7\%-3.3\% worse than the upper-bound final model accuracy (Figure~\ref{fig:breakdown-final-accuracy}).

\subsubsection{Sensitivity Analysis}
\label{eval:sensitivity}

\begin{figure}[t]
  \centering
  {
    \subfigure[ShuffleNet (Image). \label{fig:1k-shufflenet}]{\includegraphics[width=0.49\linewidth]{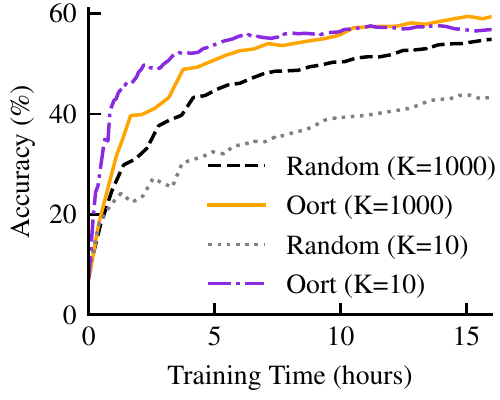}} \hfill
    \subfigure[Albert (LM). \label{fig:1k-lm}]{\includegraphics[width=0.49\linewidth]{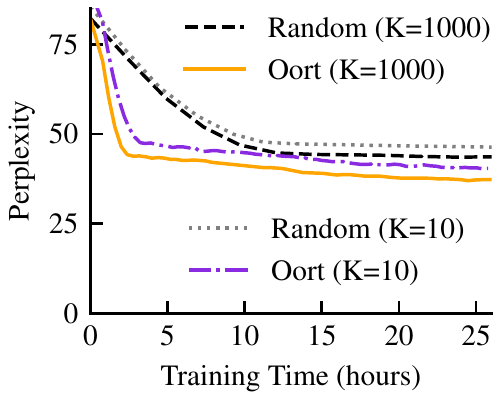}}
  }
  \caption{\name outperforms in different scales of participants.}
  \label{fig:1k-tta}
\end{figure}

\paragraph{Impact of number of participants $K$.}
We evaluate \name across different scales of participants in each round, 
where we cut off the training after 200 rounds given the diminishing rewards.
We observe that \name improves time-to-accuracy efficiency across different number of participants (Figure~\ref{fig:1k-tta}), and having more participants in FL indeed receives diminishing rewards. 
This is because taking more participants (i) is similar to having a large batch size, 
which is confirmed to be even negative to round-to-accuracy performance \cite{batchsize-iclr}; 
(ii) can lead to longer rounds due to stragglers when the number of clients is limited (\eg, $K$=1000 on OpenImage dataset). 

\begin{figure}[t]
  \centering
  {
    \subfigure[ShuffleNet (Image). \label{fig:alpha-shufflenet}]{\includegraphics[width=0.49\linewidth]{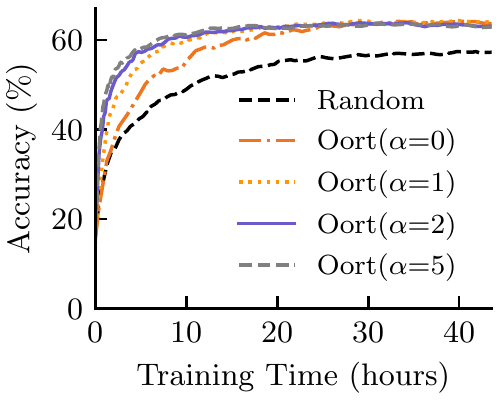}} \hfill
    \subfigure[Albert (LM). \label{fig:alpha-lm}]{\includegraphics[width=0.49\linewidth]{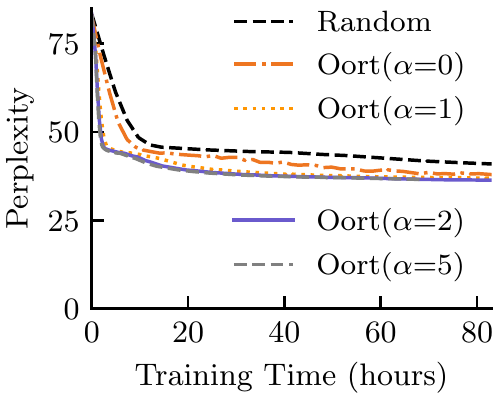}}
  }
  \caption{\name improves performance across penalty factors.}
  \label{fig:alpha-sensitivity}
\end{figure}

\paragraph{Impact of penalty factor $\alpha$ on stragglers.}
\update{
\name uses the penalty factor $\alpha$ to penalize the utility of stragglers in participant selection, whereby it adaptively prioritizes high system efficiency participants.
Figure~\ref{fig:alpha-sensitivity} shows that \name outperforms its counterparts across different $\alpha$. Note that \name orchestrates its components to automatically navigate the best performance across parameters: larger $\alpha$ (\ie, overemphasizing system efficiency) drives the Pacer to relax the system constraint $T$ more frequently to admit clients with higher statistical efficiency, and vice versa. As such, \name achieves similar performance across all non-zero $\alpha$.
}

\begin{figure}[t]
  \centering
  {
    \subfigure[Corrupted clients. \label{fig:acc-malicious_client}]{\includegraphics[width=0.49\linewidth]{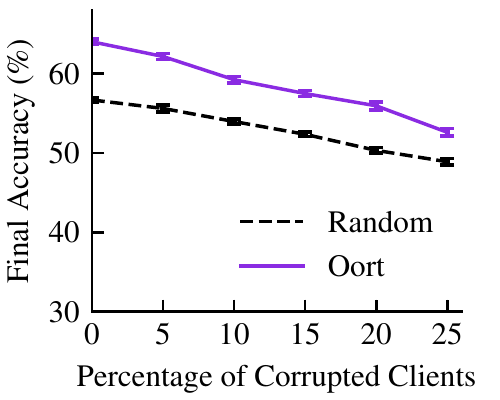}} \hfill
    \subfigure[Corrupted data. \label{fig:acc-malicious_data}]{\includegraphics[width=0.49\linewidth]{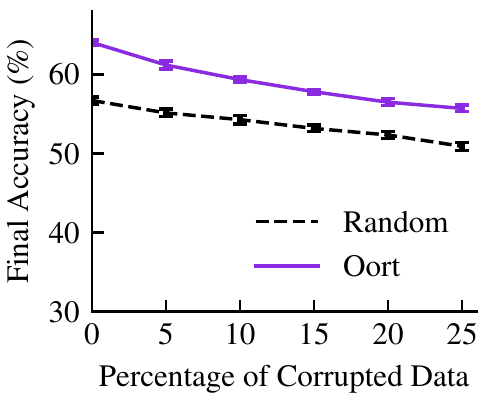}}
  }
  \caption{\name still improves performance under outliers.}
  \label{fig:noisy-data}
\end{figure}

\paragraph{Impact of outliers.}
We investigate the robustness of \name by introducing outliers manually. 
Following the popular adversarial ML setting \cite{poison-fl}, we randomly flip the ground-truth data labels of the OpenImage dataset to any other categories, resulting in artificially high utility. 
We consider two practical scenarios with the ShuffleNet model: 
(i) Corrupted clients: labels of all training samples on these clients are flipped (Figure~\ref{fig:acc-malicious_client}); 
(ii) Corrupted data: each client uniformly flips a subset of her training samples (Figure~\ref{fig:acc-malicious_data}).
We notice \name still outperforms across all degrees of corruption.

\begin{figure}[t]
  \centering
  {
    \subfigure[Round to accuracy (MobileNet). \label{fig:noise-rta-m}]{\includegraphics[width=0.49\linewidth]{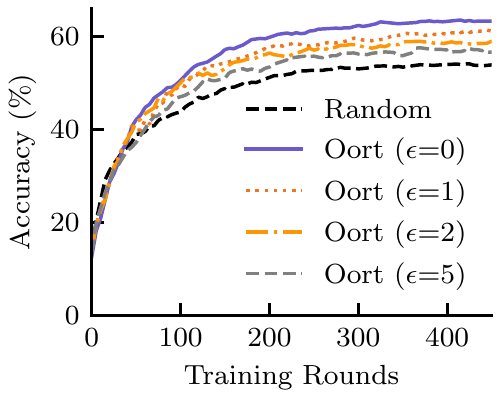}} 
    \subfigure[Time to accuracy (MobileNet). \label{fig:noise-tta-m}]{\includegraphics[width=0.49\linewidth]{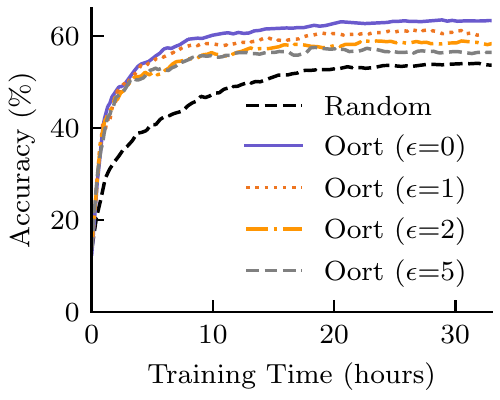}}
    \subfigure[Round to accuracy (ShuffleNet). \label{fig:noise-rta}]{\includegraphics[width=0.49\linewidth]{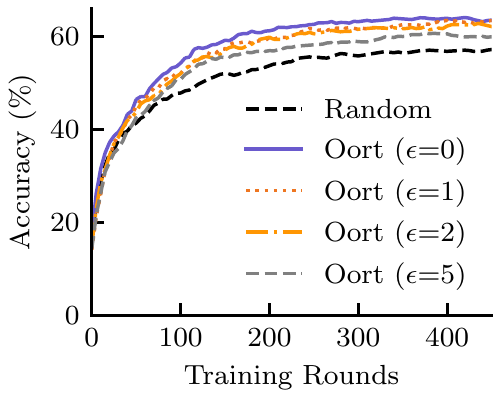}} 
    \subfigure[Time to accuracy (ShuffleNet). \label{fig:noise-tta}]{\includegraphics[width=0.49\linewidth]{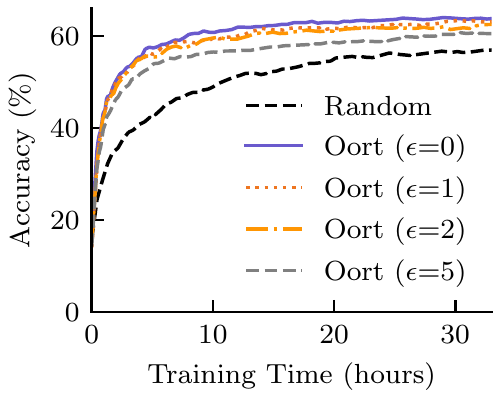}}
  }
  \caption{\name improves performance even under noise.}
  \label{fig:noise-kuiper}
\end{figure}

\paragraph{Impact of noisy utility.}
\update{
We next show the superior performance of \name over its counterparts under noisy utility value.
In this experiment, we add noise from the Gaussian distribution $Gaussian(0, \sigma^2)$, 
and investigate \name's performance with different $\sigma$. 
Similar to differential FL \cite{diff-fl}, we define $\sigma=\epsilon \times Mean(real\_value)$, 
where $Mean(real\_value)$ is the average real value without noise. 
Note that we take this $real\_value$ as reference for the ease of presentations, 
and developers can refer to other values.
As such, a large $\epsilon$ implies larger variance in noise, 
thus providing better privacy by disturbing the real value significantly.
We report the statistical efficiency after adding noise to the statistical utility (Fig~\ref{fig:noise-rta-m} and Fig~\ref{fig:noise-rta}), as well as the time-to-accuracy performance (Fig~\ref{fig:noise-tta-m} and Fig~\ref{fig:noise-tta}). We observe that \name still improves performance across different amount of noise, and is robust even when the noise is large (\eg, $\epsilon=5$ is often considered to be very large noise \cite{diff-dp}).
}

\renewcommand{\arraystretch}{1.18}
\begin{table}[t]
\centering 
\small
\setlength{\arrayrulewidth}{1pt}
\begin{tabularx}{\linewidth}{c c c c c c c c c c} 
\hhline{-|-|-|-|} 
Strategy & TTA (h)  & Final Accuracy (\%)  & Var. (Rounds) \\ 
\hhline{-|-|-|-|}
Random  & 36.3  &  57.3 &  0.39 \\
\hhline{-|-|-|-|}
$f$ = 0   & 5.8  & 64.2  &  6.52 \\
$f$ = 0.25 & 6.1  &  62.4 & 5.1 \\
$f$ = 0.5 &  13.1 & 59.7  &  2.03 \\
$f$ = 0.75 & 25.4  & 58.6  & 0.65 \\
$f$ = 1   &  30.1 & 57.2  &  0.31 \\
\hhline{-|-|-|-|}
\end{tabularx}
\caption{\name improves time to accuracy (TTA) across different fairness knobs ($f$). Random reports the performance of random participant selection. The variance of rounds reports how fairness is enforced in terms of the number of participating rounds across clients. A smaller variance implies better fairness.}
\label{table:round-fair}
\end{table}

\paragraph{\name can respect developer-preferred fairness.} 
\update{
In this experiment, we expect all clients should have participated training with the same number of rounds (Table~\ref{table:round-fair}), implying a fair resource usage \cite{fl-survey}. We train ShuffleNet model on OpenImage dataset with YoGi. 
To this end, we sweep different knobs $f$ to accommodate the developer demands for the time-to-accuracy efficiency and fairness. Namely, we replace the current utility definition of client $i$ with $(1-f)\times Util(i) + f \times fairness(i)$, where $fairness(i)= max\_resource\_usage - {resource\_usage(i)}$. 
Understandably, time-to-accuracy efficiency significantly decreases as $f\to 1$, 
since we gradually end up with round-robin participant selection, totally ignoring the utility of clients. 
Note that \name still achieves better time-to-accuracy even when $f\to 1$ as it prioritizes high system utility clients 
 at the beginning of training, thus achieving shorter rounds. 
Moreover, \name can enforce different fairness preferences while improving efficiency across fairness knobs. 
}

\subsection{FL Testing Evaluation}
\label{eval:testing}


\subsubsection{Preserving Data Representativeness}
\label{eval:non-clair}
\begin{figure}[t]
  \centering
  {
    \subfigure[Google Speech. \label{fig:nc-speech}]{\includegraphics[width=0.49\linewidth]{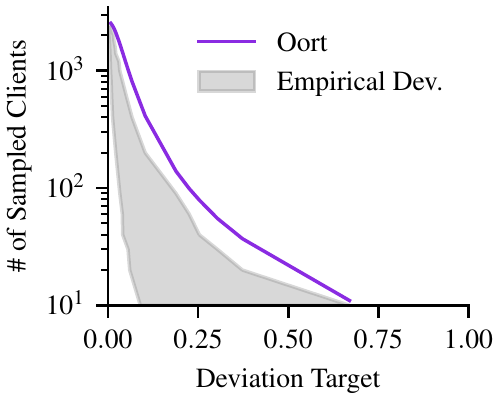}} 
    \subfigure[Reddit. \label{fig:nc-so}]{\includegraphics[width=0.49\linewidth]{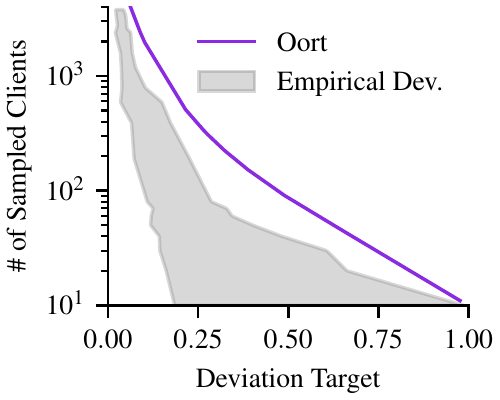}}
  }
  \caption{\name can cap data deviation for all targets. Shadow indicates the empirical [min, max] range of the x-axis values over 1000 runs given the y-axis input.}
  \label{fig:testing-nc}
\end{figure}

\paragraph{\name can cap data deviation.} 
Figure~\ref{fig:testing-nc} reports \name's performance on serving different deviation targets, with respect to the global distribution. 
We sweep the number of selected clients from 10 to 4k, 
and randomly select each given number of participants over 1k times to empirically search their possible deviation. 
We notice that for a given deviation target, 
(i) different workloads require distinct number of participants. For example, to meet the target of 0.05 divergence, the Speech dataset uses 6$\times$ less participants than the Reddit attributing to its smaller heterogeneity (\eg, tighter range of the number of samples); 
(ii) with the \name-determined number of participants, no empirical deviation exceeds the target, showing the effectiveness of \name in satisfying the deviation target, whereby 
\name reduces the cost of expanding participant set arbitrarily and improves the testing duration. 

\subsubsection{Enforcing Diverse Data Distribution}
\label{eval:testing-clair}



\begin{figure}[t]
  \centering
  {
    \subfigure[OpenImage (Testing duration). \label{fig:e2e-openimg}]{\includegraphics[width=0.49\linewidth]{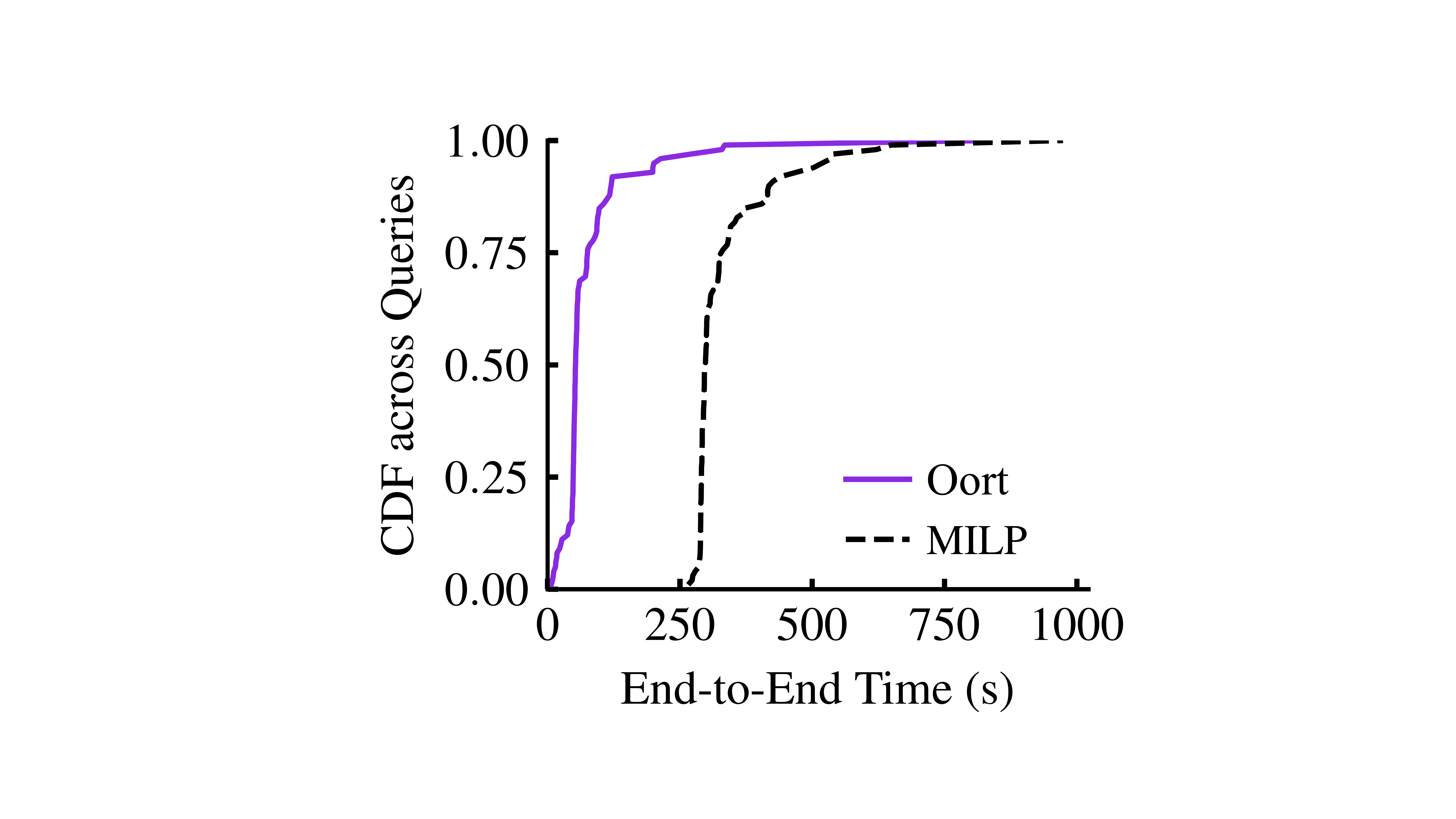}} 
    \subfigure[OpenImage (Overhead). \label{fig:overhead-openimg}]{\includegraphics[width=0.49\linewidth]{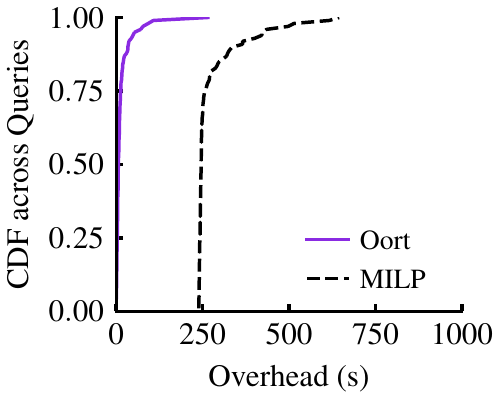}}
  }
  \caption{\name outperforms MILP in clairvoyant FL testing.}
  \label{fig:testing-lp}
\end{figure}

\paragraph{\name outperforms MILP.} 


We start with the middle-scale OpenImage dataset and compare the end-to-end testing duration of \name and MILP. 
Here, we generate 200 queries using the form ``Give me \textit{X} representative samples'',  
where we sweep \textit{X} from 4k to 200k and budget \textit{B} from 100 participants to 5k participants. 
We report the validation time of MobileNet on participants selected by these strategies. 

Figure~\ref{fig:e2e-openimg} shows the end-to-end testing duration.
We observe \name outperforms MILP by 4.7$\times$ on average. 
This is because \name suffers little computation overhead by greedily reducing the search space of MILP. 
As shown in Figure~\ref{fig:overhead-openimg}, MILP takes 274 seconds on average to complete the participant selection, 
while \name only takes 15 seconds. 



\begin{figure}[t]
  \centering
  {
    \subfigure[StackOverflow (0.3M clients).]{\includegraphics[width=0.49\linewidth]{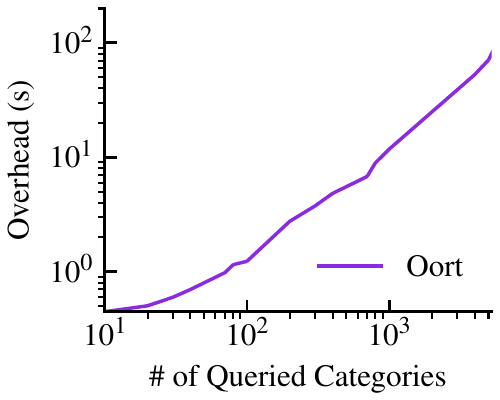}} 
    \subfigure[Reddit (1.6M clients).]{\includegraphics[width=0.49\linewidth]{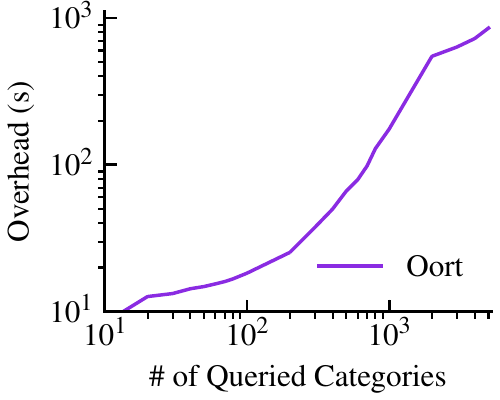}}
  }
  \caption{\name scales to millions of clients, while MILP did not complete on any query.}
  \label{fig:test-macro}
\end{figure}

\paragraph{\name is scalable.} 
\update{
We further investigate \name's performance on the large-scale StackOverflow and Reddit dataset with millions of clients, 
where we take 1\% of the global data as the requirement, 
and sweep the number of interested categories from 1 to 5k. 
Figure~\ref{fig:test-macro} shows even though we gradually magnify the search space of participant selection by introducing more categories, 
\name can serve our requirement in a few minutes at the scale of millions of clients, while MILP fails to generate the solution decision for any query. 
}



\section{Related Work}
\label{sec:related}

\paragraph{Federated Learning}
Federated learning \cite{fl-survey} is a distributed machine learning paradigm 
in a network of end devices, 
wherein Prox\cite{fl-heternet} and YoGi\cite{fed-yogi} are state-of-the-art optimizations in tackling data heterogeneity.
Recent efforts in FL have been focusing on improving communication efficiency \cite{fedavg, gaia-cmu} or compression schemes \cite{qsgd}, 
ensuring privacy by leveraging multi-party computation (MPC) \cite{smc} and differential privacy \cite{diff-fl}, or tackling heterogeneity by reinventing ML algorithms \cite{adapt-sgd, fl-fair}. 
However, they underperform in FL because of the suboptimal participant selection they rely on,  
and lack systems supports for developers to specify their participant selection criteria.


\paragraph{Datacenter Machine Learning}
Distributed ML in datacenters has been well-studied \cite{bytescheduler, pipedream, taso}, 
wherein they assume relatively homogeneous data and workers \cite{themis, tiresia}. 
While developer requirements and models can still be the same, 
the heterogeneity of client system performance and data distribution makes FL much more challenging. We aim at enabling them in FL.
\update{
To accelerate traditional model training, some techniques bring up importance sampling to prioritize important training samples in selecting mini-batches for training \cite{is-icml, is-nips, importance-sampling}. 
While bearing some resemblance in prioritizing data, 
\name adaptively considers both statistical and system efficiency in formulating the client utility at scale.
}

\paragraph{Geo-distributed Data Analytics}
\update{
Federated data analytics has been a topic of interest in geo-distributed storage \cite{pando} and data processing systems \cite{awstream, sol} that attempt to reduce latency \cite{clarinet} and/or save bandwidth \cite{iridium, geode, relay-hotcloud}. 
Gaia \cite{gaia-cmu} reduces network traffics for model training across datacenters, while Sol \cite{sol} enables generic federated computation on data with sub-second latency in the execution layer.
These work back up \name with cross-layer system support, whereas \name cherry-picks participants before execution.
}

\paragraph{Privacy-preserving Data Analytics}
To gather sensitive statistics from user devices,
several differentially private systems add noise to user inputs locally to ensure privacy \cite{rappor}, but this can reduce the accuracy. 
Some assume a trusted third party, which only adds noise to the aggregated raw inputs \cite{prochlo}, or use MPC to enable global differential privacy without a trusted party \cite{honeycrisp}. 
\update{While our goal is not to address the security and privacy issue in these solutions, \name enables informed participant selection by leveraging the information already available in today’s FL, and can reconcile with them (\eg, to deliver improvement under outliers while respecting privacy).
}

\section{Conclusion}

\update{While today's FL efforts have been optimizing the statistical model and system efficiency by reinventing traditional ML designs, 
the participant selection mechanisms they rely on underperform for federated training and testing, 
and fail to enforce diverse data selection criteria. 
In this paper, we present \name to enable guided participant selection for FL developers. 
Compared to existing mechanisms, 
\name achieves large speedups in time-to-accuracy performance for federated training by picking clients with high statistical and system utility, 
and it allows developers to specify their selection criteria on data while efficiently serving their requirements on data distribution during testing even at the scale of millions of clients. 
The artifacts of \name are available at \url{https://github.com/SymbioticLab/Oort}.
}



\section*{Acknowledgments}
\label{sec:ack}
\update{Special thanks go to the entire ConFlux team and CloudLab team for making \name experiments possible. We would also like to
thank the anonymous reviewers, our shepherd, Gennady Pekhimenko, and SymbioticLab members for their insightful feedback. This work was supported in part by NSF grants CNS-1900665 and CNS-1909067.}

\label{EndOfPaper}

{
\bibliographystyle{plain}
\bibliography{kuiper}
}
\clearpage
\appendix

\section{Proving Benefits of Statistical Utility}
\label{app:utility-proof}

We follow the proof of importance sampling to show the advantage of our statistical utility in theory. 
%
The convergence speed of Stochastic Gradient Descent (SGD) can be defined as the reduction $R$ of the divergence of model weight $\bold{w}$ from its optimal $\bold{w^*}$ in two consecutive round $t$ and $t+1$ \cite{importance-sampling, is-icml} :
\begin{align}
R &= \mathbbm{E}\big[\parallel \bold{w_{t}} - \bold{w^*}\parallel_{2}^{2} - \parallel \bold{w_{t+1}} - \bold{w^*}\parallel_{2}^{2} \big]
\label{eq:speed-reduction}
\end{align}

\paragraph{How does oracle sampling help in theory?} 
If the learning rate of SGD is $\eta$ and we use loss function $L$ to measure the training loss between input features $x$ and the label $y$, then $\bold{w_{t+1}}$ = $\bold{w_t} - \eta \nabla L(\bold{w_t}(x_i), y_i)$. We now set the gradient $G_t=\nabla L(\bold{w_t}(x_i), y_i)$ for brevity, then from Eq.~(\ref{eq:speed-reduction}):
\begin{align}
R &= -\mathbbm{E}\big[(\bold{w_{t+1}} - \bold{w^*})^{T}(\bold{w_{t+1}} - \bold{w^*}) - (\bold{w_{t}} - \bold{w^*})^{T}(\bold{w_{t}} - \bold{w^*}) \big] \nonumber \\ 
&= -\mathbbm{E}\big[ \bold{w_{t+1}^T}\bold{w_{t+1}} - 2\bold{w_{t+1}}\bold{w^*} - \bold{w_{t}^T}\bold{w_t} + 2\bold{w_{t}}\bold{w^*}\big] \nonumber \\ 
&= -\mathbbm{E}\big[ (\bold{w_t - \eta G_t})^T(\bold{w_t - \eta G_t}) + 2\eta G_t^T\bold{w^*} - \bold{w_t}^T\bold{w_t} \big] \nonumber \\ 
&= -\mathbbm{E}\big[ -2\eta(\bold{w_t}-\bold{w^*})G_t + \eta^2 G_t^T G_t\big] \nonumber \\ 
&= \small{2\eta(\bold{w_t}-\bold{w^*})\mathbbm{E}[G_t] - \eta^2\mathbbm{E}[G_t]^T\mathbbm{E}[G_t] - \eta^2Tr(\mathbbm{V}[G_t]) 
\label{eq:variance-reduction}
}
\end{align}

It has been proved that optimizing the first two terms of Eq.~(\ref{eq:variance-reduction}) is intractable due to their joint dependency on $\mathbbm{E}[G_t]$, however, one can gain a speedup over random sampling by intelligently sampling important data bins to minimize $Tr(\mathbbm{V}[G_t])$ (\ie, reducing the variance of gradients while respecting the same expectation $\mathbbm{E}[G_t]$) \cite{is-nips, importance-sampling}. Here, the oracle is to pick bin $B_i$ with a probability proportional to its importance $|B_i| \sqrt{\frac{1}{|B_i|}\sum_{k \in B_i} \parallel G(k) \parallel^2}$, where $\parallel G(k) \parallel$ is the L2-norm of the unique sample $k$'s gradient $G(k)$ in bin $B_i$ (Please refer to \cite{bin-sampling} for detailed proof). 

\paragraph{How does loss-based approximation help?}
We have shown the advantage of importance sampling by sampling the larger gradient norm data, and next we present the theoretical insights  that motivates our loss-based utility design. 
\begin{corollary}
(Theorem 2 in \cite{loss-sampling}). Let $\parallel G(k) \parallel$ denote the gradient norm of any sample $k$, $M=\max{\parallel G(k) \parallel}$. There exists $K>0$ and $C<M$ such that $\frac{1}{K} L(\bold{w}(x_k), y_k) + C \geq \parallel G(k) \parallel$.
\end{corollary}

This corollary implies that a bigger loss leads to a large upper bound of the gradient norm. 
To sample data with a larger gradient norm, we prefer to pick the one with bigger loss.
Moreover, it has been empirically shown that sampling high loss samples exhibits similar variance reducing properties to sampling according to the gradient norm, resulting in better convergence speed compared to naive random sampling \cite{importance-sampling}. 
Readers can refer to the recent FL work~\cite{selection-loss} for more proof.

By taking account of the oracle and the effectiveness of loss-based approximation, we propose our loss-based statistical utility design, whereby we achieve the close to upper-bound statistical performance (\S\ref{eval:eval-breakdown}).

\section{Privacy Concern in Collecting Feedbacks}
\label{app:privacy}

Depending on different requirements on privacy, 
we elaborate on how \name respects privacy while outperforming existing mechanisms (\S\ref{sec:stats-util}).

\paragraph{Compute aggregated training loss on clients locally.}
Our statistical utility of a client relies on the aggregated training loss of all samples on that client. 
The training loss of each sample measures the prediction uncertainties of model on every possible output (\eg, category), 
and even the one with a correct prediction can generate non-zero training loss \cite{testing-bias}.
So it does not reveal raw data inputs. 
Moreover, it does not leak the categorical distribution either, 
since samples of the same category can own different training losses due to their heterogeneous input features. 
Note that even when we consider homogeneous input features, 
a high total loss can be from several high loss samples, or many moderate loss samples.

\paragraph{Add noise to hide the real aggregated loss.} 
Even when clients have very stringent privacy concerns on their aggregated loss, 
clients can add noise to the exact value. 
Similar to the popular differentially private FL \cite{diff-fl}, 
clients can disturb their real aggregated loss by adding Gaussian noise (\ie, noise from the Gaussian distribution).
In fact, \name can tolerate this noisy statistical utility well, 
owing to its probabilistic selection from the pool of high-utility clients, 
wherein teasing apart the top-k\% utility clients from the rest is the key. 

We first prove that \name is still very likely to select high-utility clients even in the presence of noise.
To pick $K$ participants out of $N$ all feasible clients, there are totally $\tbinom{N}{K}$ possible combinations. 
We denote these combinations as $X_i$ and sort them $X_1 \leq  \ldots \leq X_n$ by the ascending order of total utility. 
Adding noise to each client ends up with an accumulated noise on $X_i$. 
Thereafter, $X_i$ turns to random variables $\bold{X}_i$ that follow the distribution of accumulated noise. 
Specifically, distribution of $\bold{X}_i$ is equivalent to shifting the distribution of noise horizontally by a constant $X_i$.
Given that noise added to $X_i$ follows the same distribution, 
(i) $\bold{X}_i$ experiences the same standard deviation for every $i$;  
(ii) the expectation of $\bold{X}_i$ is the sum of $X_i$ and the expectation of noise. 
Note that adding a constant (\ie, the expectation of noise here) to the inequality does not change its properties, 
so we still have $\mathbbm{E}[\bold{X}_1] \leq  \ldots \leq \mathbbm{E}[\bold{X}_n]$. 
As such, we are more likely to select high-utility clients (\ie, combination $\bold{X}_i$ with higher $\mathbbm{E}[\bold{X}_i]$) in sampling when picking $i$ with the highest value of $X_i$. 
Moreover, we have show the superior empirical performance of \name over its counterparts under noise in Section~\ref{eval:sensitivity}.


\paragraph{Rely on gradient norm of batches.}
For case where clients are even reluctant to report the noisy loss, 
we introduce an alternative statistical utility to drive our exploration-exploitation based client selection.
Our intuition is to use the gradient norm of batches to approximate the gradient norm of individual samples $\nabla f(k)$ in the oracle importance $|B_i| \sqrt{\frac{1}{|B_i|}\sum_{k \in B_i} \parallel \nabla f(k) \parallel^2}$.
In mini-batch SGD, we have   
\begin{align}
\vspace{-.4cm}
\bold{w}_{t+1} = \bold{w}_{t} - learning\_rate \times \frac{1}{batch\_size} \times \sum\limits_{k \in batch} \nabla f(k)\nonumber
\vspace{-.4cm}
\end{align}
where $w_{t}$ is the model weights at time $t$.
Now, we can use the gradient norm of batches (\ie, $\parallel \bold{w}_{t+1} - \bold{w}_{t} \parallel^2$) to approximate $\parallel \nabla f(k) \parallel^2$, and they become equivalent when the batch size is 1.
Note that today's FL is already collecting the model updates (\ie, $\bold{w}_{t+1} - \bold{w}_{t}$), 
so we are not introducing additional information.
As such, we consider the client with larger accumulated gradient norm of batches to be more important. 

\begin{figure}[t]
  \centering
  {
    \subfigure[MobileNet. \label{fig:norm-mobilenet}]{\includegraphics[width=0.49\linewidth]{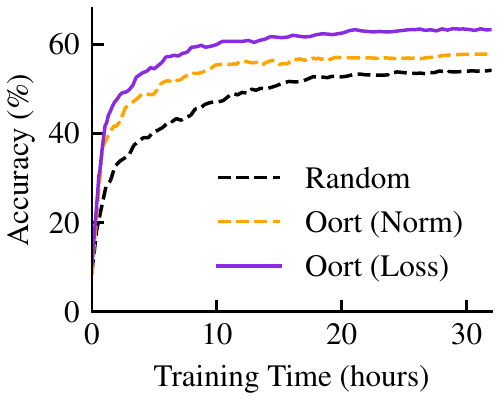}} 
    \subfigure[ShuffleNet. \label{fig:norm-shufflenet}]{\includegraphics[width=0.49\linewidth]{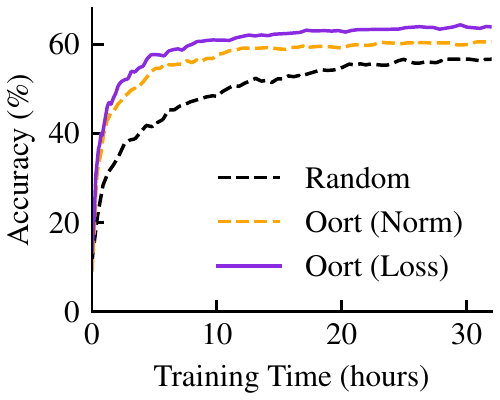}}
  }
  \caption{\name outperforms with different utility definitions.}
  \label{fig:norm-tta}
\end{figure}

We report the empirical performance of this approximation and the loss-based statistical utility using YoGi. 
As shown in Fig~\ref{fig:norm-tta}, \name achieves superior performance over the random selection, and the loss-based utility is better than its counterparts. 
This is because the approximation accuracy with the norm of batches decreases when using mini-batch SGD, whereas 
mini-batch SGD is more popular than the single-sample batch in ML.

\section{Determining Size of Participants}
\label{app:group-size}

We next introduce Lemma~\ref{lemma:group-size}, which captures how the empirical value of $\bar{X}$ (\ie, average number of samples of participants for category $X$) deviates from the expectation $E[\bar{X}]$ (\ie, average number of samples of all clients) as the size of participants $n$ varies. 

%
\begin{lemma}
\label{lemma:group-size}
For a given tolerance on deviation $\epsilon$ and confidence interval $\delta$ for category $X$, the number of participants $n$ we need to achieve $Pr[|\bar{X}-{E}[\bar{X}]| < \epsilon] > \delta$ requires: 
\begin{align}
n \geq (N+1) \times \frac{1}{1-\frac{2N}{\log (1-\delta)} \times (\frac{\epsilon}{\max\{X\}-\min\{X\}})^2}
\label{equ:lower-bound}
\end{align}
where $N$ is the total number of feasible clients, and $\max\{X\}$ and $\min\{X\}$ denote the global maximum and minimum possible number of samples that all clients can hold, respectively.
\end{lemma}
Lemma~\ref{lemma:group-size} is a corollary of Hoeffding-Serfling Bound \cite{concentration}, and we omit the detailed proof for brevity. 
Intuitively, when we have an extremely stringent requirement (\ie, $\epsilon \to 0$), we have to include more participants (\ie, $n\to N$).
When more information of the client data characteristics is available, one can refine this range better. For example, the bound of Eq. (\ref{equ:lower-bound}) can be improved with Chernoff’s inequality \cite{concentration} when the distribution of sample quantities is provided.

Similarly, the multi-category scenario proves to be an instance of multi-variate Hoeffding Bound.
Given the developer-specific requirement on each category, 
the developer may want to figure out how many participants needed to satisfy all these requirements simultaneously 
(\eg, $Pr[|\bar{X}-{E}[\bar{X}] | < \epsilon{_x} \wedge |\bar{Y}-{E}[\bar{Y}] | < \epsilon{_y}] > \delta$).  
More discussions are out of the scope of this paper, but readers can refer to \cite{mutivariate-hoeff} for detailed discussions and a complete solution.

\end{document}